\documentclass{article}

\PassOptionsToPackage{numbers,sort&compress}{natbib}

\usepackage[final]{neurips_2023}
\usepackage{hypernat}

\usepackage[utf8]{inputenc} %
\usepackage[T1]{fontenc}    %
\usepackage{url}            %
\usepackage{booktabs}       %
\usepackage{amsfonts}       %
\usepackage{enumitem}       %
\usepackage{float}       %
\usepackage{nicefrac}       %
\usepackage{microtype}      %
\usepackage[table]{xcolor}
\usepackage[citecolor=blue,pagebackref=true,breaklinks=true,letterpaper=true,colorlinks,bookmarks=false]{hyperref}
\usepackage{adjustbox}
\usepackage{graphicx}
\usepackage{amsmath}
\usepackage{amssymb}
\usepackage{amsthm}
\usepackage{bm}
\usepackage{pdfpages}
\usepackage{booktabs}
\usepackage[font=smaller]{caption}
\usepackage{subcaption}
\usepackage{wrapfig}  

\newcommand*{\rev}{\textcolor{black}}
\newcommand{\VS}{\textcolor{black}}
\newcommand{\AT}{\textcolor{black}}

\newcommand{\tianwei}{\textcolor{blue}}

\theoremstyle{definition}
\newtheorem{proposition}{Proposition}

\usepackage{amsmath,amsfonts,bm}

\def\eqref#1{equation~\ref{#1}}

\def\1{\bm{1}}

\definecolor{forward}{RGB}{117,33,33}
\def\forwardmodel{\textcolor{forward}{\texttt{forward}}}
\def\warp{\textcolor{forward}{\texttt{warp}}}
\def\render{\textcolor{forward}{\texttt{render}}}
\def\generate{\textcolor{forward}{\texttt{synthesize}}}
\def\denoise{{\texttt{denoise}}}
\def\sig{{\rmS}}
\def\obs{{\rmO}}
\def\param{{\phi}}
\def\context{\rmO^\text{ctxt}}
\def\target{\rmO^\text{trgt}}

\def\targetnovel{\rmO^\text{novel}}
\def\targett{\rmO^\text{trgt}_t}
\def\targetdet{\rmO^\text{trgt}_\text{det}}

\def\paramcontext{{\phi^\text{ctxt}}}
\def\paramtarget{{\phi^\text{trgt}}}

\def\paramtargetnovel{{\phi^\text{novel}}}
\def\signalestimate{{\sig_{t\text{-}1}}}
\def\targetestimate{\hat{\obs}^\text{trgt}_{t\text{-}1}}
\def\targetestimatenovel{\hat{\obs}^\text{novel}_{t\text{-}1}}

\def\rvc{{\mathbf{c}}}

\def\rvu{{\mathbf{i}}}

\def\rvp{{\mathbf{p}}}

\def\rvu{{\mathbf{u}}}

\def\rvx{{\mathbf{x}}}

\def\rmO{{\mathbf{O}}}

\def\rmS{{\mathbf{S}}}

\DeclareMathAlphabet{\mathsfit}{\encodingdefault}{\sfdefault}{m}{sl}
\SetMathAlphabet{\mathsfit}{bold}{\encodingdefault}{\sfdefault}{bx}{n}

\newcommand{\feat}{\mathbf{F}}
\newcommand{\enc}{\texttt{enc}}
\newcommand{\enct}{\texttt{enc}_t}
\newcommand{\MLP}{\texttt{MLP}}

\def\xpix{\mathbf{p}^\text{pix}}
\def\point{\mathbf{p}}

\def\Feat{\mathbf{F}}
\def\feat{\mathbf{f}}

\title{Diffusion with Forward Models: Solving Stochastic Inverse Problems Without Direct Supervision}

\author{%
        Ayush Tewari\textsuperscript{1$^*$}\qquad
	Tianwei Yin\textsuperscript{1$^*$}\qquad
	George Cazenavette\textsuperscript{1}\qquad
	Semon Rezchikov\textsuperscript{4}\qquad \\
        \textbf{Joshua B. Tenenbaum}\textsuperscript{1,2,3}\qquad
	\textbf{Fr\'{e}do Durand}\textsuperscript{1}\qquad
 	\textbf{William T. Freeman}\textsuperscript{1}\qquad
	\textbf{Vincent Sitzmann}\textsuperscript{1}
}

\begin{document}

\maketitle
\vbox{%
	\vskip -0.26in
	\hsize\textwidth
	\linewidth\hsize
	\centering
	\normalsize
	\footnotemark[1]MIT CSAIL \quad \footnotemark[2]MIT BCS \quad \footnotemark[3]MIT CBMM \quad \footnotemark[4]Princeton IAS
	\vskip 0.2in
}
\let\thefootnote\relax\footnotetext{$^{*}$ Equal Contribution. Project page: \url{diffusion-with-forward-models.github.io}}

\begin{abstract}
Denoising diffusion models have emerged as a powerful class of generative models capable of capturing the distributions of complex, real-world signals. 
However, current approaches can only model distributions for which training samples are directly accessible, which is not the case in many real-world tasks. 
In inverse graphics, for instance, we seek to sample from a distribution over 3D scenes consistent with an image but do not have access to ground-truth 3D scenes, only 2D images.
We present a new class of \VS{conditional} denoising diffusion probabilistic models that learn to sample from distributions of signals that are never observed directly, but instead are only measured through a known differentiable forward model that generates partial observations of the unknown signal.
To accomplish this, we directly integrate the forward model into the denoising process.
At test time, our approach enables us to sample from the distribution over underlying signals consistent with some partial observation. 
We demonstrate the efficacy of our approach on three challenging computer vision tasks. For instance, in inverse graphics, we demonstrate that our model \VS{in combination with a 3D-structured conditioning method} enables us to directly sample from the distribution of 3D scenes consistent with a single 2D input image.
\end{abstract}

\section{Introduction}
Consider the problem of reconstructing a 3D scene from a single picture. 
Since much of the 3D scene is unobserved, there are an infinite number of 3D scenes that could have produced the image, due to the 3D-to-2D projection, occlusion, and limited field-of-view that leaves a large part of the 3D scene unobserved. 
Given the ill-posedness of this problem, it is desirable for a reconstruction algorithm to be able to sample from the distribution over all plausible 3D scenes that are consistent with the 2D image, generating unseen parts in plausible manners.
Previous data-completion methods, such as in-painting in 2D images, are trained on large sets of ground-truth output images along with their incomplete (input) counterparts. %
Such techniques do not easily extend to 3D scene completion, since curating a large dataset of ground-truth 3D scene representations is very challenging.

This 3D scene completion problem, known as inverse graphics, is just one instance of a broad class of problems often referred to as \emph{Stochastic Inverse Problems}, 
which arise across scientific disciplines whenever we capture partial observations of the world through a sensor.
In this paper, we introduce a diffusion-based framework that can tackle this problem class, enabling us to sample from a distribution of signals that are consistent with a set of partial observations that are generated from the signal by a non-invertible, generally nonlinear, forward model. For instance, in inverse graphics, we learn to sample 3D scenes given an image, yet never observe paired observations of images and 3D scenes at training time, nor observe 3D scenes directly.

While progress in deep learning for generative modeling has been impressive, this problem remains unsolved.
In particular, variational autoencoders and conditional neural processes are natural approaches but have empirically fallen short of modeling the multi-modal distributions required in, for instance, inverse graphics. They have so far been limited to simple datasets. 
Emerging diffusion models~\cite{sohl2015deep}, in contrast, enable sampling from highly complex conditional distributions but require samples from the output distribution that is to be modeled for training, e.g. full 3D models.
Some recent work in inverse graphics has resorted to a two-stage approach, where one first reconstructs a large dataset of 3D scenes to then train an image-conditional diffusion model to sample from the conditional distribution over these scenes~\cite{mueller2022diffrf, kim2023neuralfield}.
To avoid a two-stage approach, another recent line of work trains a conditional diffusion model to sample from the distribution over novel views of a scene, only requiring image observations at training time~\cite{chan2023generative,zhou2022sparsefusion}.
However, such methods do \emph{not} model the distribution over 3D scenes directly and therefore cannot sample from the distribution over 3D scenes consistent with an image observation. Thus, a multi-view consistent 3D scene can only be obtained in a costly post-processing stage\tianwei{~\cite{poole2022dreamfusion}}. 
\VS{A notable exception is the recently proposed RenderDiffusion~\cite{anciukevivcius2022renderdiffusion}, demonstrating that it is possible to train an unconditional diffusion model over 3D scenes from observing only monocular images.  While one can perform conditional sampling even with unconditional models, they are fundamentally limited to simple distributions, in this case, single objects in canonical orientations.}

\VS{Our core contribution is a novel approach for integrating any differentiable forward model that describes how partial observations are obtained from signals, such as 2D image observations and 3D scenes, with conditional denoising diffusion models.}
By sampling an observation from our model, we jointly sample the signal that gave rise to that observation. 
Our approach has a number of advantages that make it highly attractive for solving complex Stochastic Inverse Problems. First, our model is trained end-to-end and does away with two-stage approaches that first require reconstruction of a large dataset of signals. Second, our model directly yields diverse samples of the signal of interest. For instance, in the inverse graphics setting, our model directly yields highly diverse samples of 3D scenes consistent with an observation that can then be rendered from novel views with guaranteed multi-view consistency. Finally, our model naturally leverages domain knowledge in the form of known forward models, such as differentiable rendering, with all guarantees that such forward models provide.
We validate our approach on three challenging computer vision tasks: inverse graphics (the focus of this paper), as well as single-image motion prediction and GAN inversion.

In summary, we make the following contributions:
\begin{enumerate}[leftmargin=1em]
    \item \looseness=-1 We propose a new method that integrates differentiable forward models with conditional diffusion models, replacing prior two-step approaches with a conditional generative model trained end-to-end.
    \item We apply our framework to build the first conditional diffusion model that learns to sample from the distribution of 3D scenes trained only on 2D images. In contrast to prior work, we \emph{directly} learn image-conditional 3D radiance field generation, instead of sampling from the distribution of novel views conditioned on a context view. \VS{Our treatment of inverse graphics exceeds a mere application of the proposed framework, contributing a novel, 3D-structured denoising step that leverages differentiable rendering both for conditioning and for the differentiable forward model.}
    \item We formally prove that under natural assumptions, as the number of observations of each signal in the training set goes to infinity, the proposed model maximizes not only the likelihood of observations, but also the likelihood of the unobserved signals.
    \item We demonstrate the efficacy of our model for two more downstream tasks with structured forward models: single-image motion prediction, where the forward model is a warping operation,  and GAN inversion, where the forward model is a pretrained StyleGAN \cite{sg2} generator.
\end{enumerate}

\section{Method}
\begin{figure}[t]
\centering
\includegraphics[width=\linewidth]{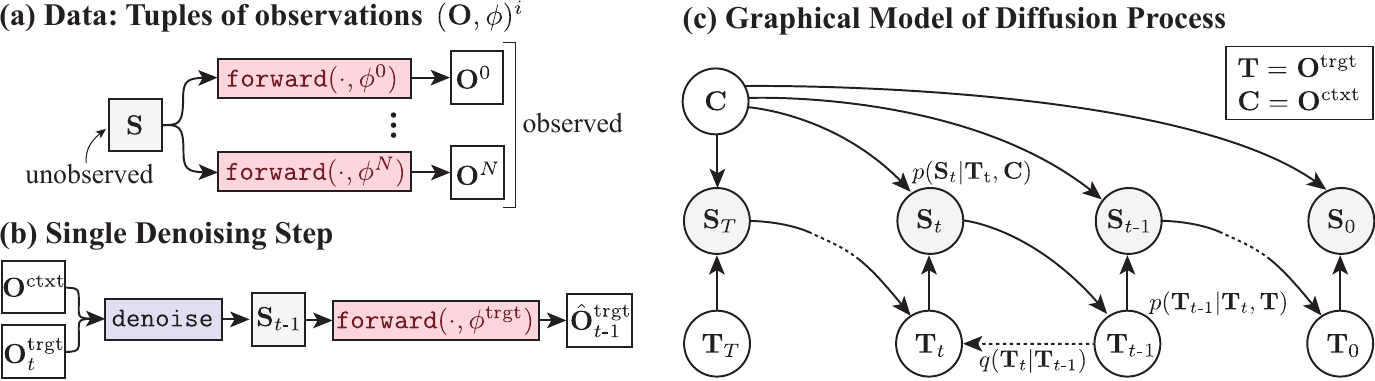}
\caption{\textbf{Overview of our proposed method.} \textbf{(a)} We assume a dataset of tuples of observations $(\obs, \param)^i$, generated from \emph{unobserved} signals $\sig$ via a differentiable forward model. \textbf{(b)} We propose to integrate the forward model directly into the denoising step of a diffusion model: given a pair of observations of the same signal, we designate context $\context$ and target $\target$. We add noise to $\target$, then feed $(\context, \paramcontext, \target_t, \paramtarget)$ to a neural network $\denoise$ to estimate the signal $\signalestimate$. We then apply the forward model to obtain an estimate of the clean target observation, $\targetestimate$. \textbf{(c)} The graphical model of the diffusion process.
\vspace{-0.45cm}
}
\label{fig:graphical}
\end{figure}
Consider observations $(\obs_j^i, \phi_j^i)$ that are generated from underlying signals $\sig_j$ according to a known forward model $\forwardmodel()$, i.e.,  $\obs_j^i = \forwardmodel(\sig_j, \param_j^i)$, where $\param_j^i$ are parameters of the forward model corresponding to observation $\obs_j^i$. 
Each observation can be \emph{partial}. Specifically, given a \emph{single} observation, there is an infinite number of signals that could have generated this observation. However, we assume that given a hypothetical set of \emph{all possible} observations, the signal is fully determined.
In the case of inverse graphics, $\obs_j^i$ are image observations of 3D scenes $\sig_j$ and $\param_j^i$ are the camera parameters, where we index scenes with $j$ and observations of the $j$-th scene via $i$. $\forwardmodel()$ is the rendering function. Note that if we were to capture \emph{every possible image} of a 3D scene, the 3D scene is uniquely determined, but given a \emph{single} image, there are an infinite number of 3D scenes that could have generated that image, both due to the fact that rendering is a projection from 3D and 2D, and due to the fact that a single image only constrains the visible part of the 3D scene.
We will drop the subscript $j$ in the following, and leave it implied that we always consider \emph{many} observations generated from \emph{many} signals. Fig.~\ref{fig:graphical} provides an illustration of the data.

We are now interested in training a model that, at test time, allows us to sample from the distribution of signals that are consistent with a previously unseen observation $\mathbf{O}$.
Formally, we aim to model the conditional distribution $p(\sig|\obs, \phi)$. %
We make the following assumptions:
\begin{itemize}[leftmargin=1em, nosep]
    \item We have access to a differentiable implementation of $\forwardmodel()$.
    \item We have access to a large dataset of observations and corresponding parameters of the forward model, $\{(\obs^i, \phi^i)\}_{i}^N$.
    \item In our training set, we have access to \emph{several} observations per signal.
\end{itemize}
Crucially, we do \emph{not} assume that we have direct access to the underlying signal that gave rise to a particular observation, i.e., we do \emph{not} assume access to tuples of $(\obs, \param, \sig)$. Further, we also do \emph{not} assume that we have access to any prior distribution over the signal of interest, i.e., we never observe a dataset of signals of the form $\{\sig^j \}_{j}$, and thus cannot train a generative model to sample from an unconditional distribution over signals.

Recent advances in deep-learning-based generative modeling have seen the emergence of denoising diffusion models as powerful generative models that can be trained to generate highly diverse samples from complex, multi-modal distributions.
We are thus motivated to leverage denoising diffusion probabilistic models to model $p(\sig|\obs, \phi)$.
However, existing approaches cannot be trained if we do not have access to signals $\sig$. In the following, we give background on denoising diffusion models and discuss the limitation.

\subsection{Background: Denoising Diffusion Probabilistic Models and their Limitation}
Denoising diffusion probablistic models are a class of generative models that learn to sample from a distribution by learning to iteratively denoise samples. 
Consider the problem of modeling the distribution $p_\theta(\rvx)$ over samples $\rvx$. 
A forward Markovian process $q(\rvx_{0:T})$ adds noise to the data as 
\begin{align}
    q(\rvx_t \mid \rvx_{t-1}) &= \mathcal{N}(\rvx_t; \sqrt{1-\beta_t} \rvx_{t-1}, \beta_t \mathbf{I}).
    \label{eq:forward}
\end{align}
Here, $\beta_t$, $t \in {1\dots T}$ are the hyperparameters that control the variance schedule. 
A denoising diffusion model learns the reverse process, where samples from a distribution $p(x_T) = \mathcal{N}(\mathbf{0, \mathbf{I}})$ are transformed incrementally into the data manifold as $p_\theta(\rvx_{0:T}) = p(x_T) \prod_{t=1}^T p_\theta(\rvx_{t-1} \mid \rvx_t)$, where 
\begin{align}
p_\theta(\rvx_{t-1} \mid \rvx_t) &= \mathcal{N}(\rvx_{t-1}; \mu(\rvx_t, t), \Sigma (\rvx_t, t)).
\label{eq:trad_denoise}
\end{align}
A neural network $\denoise_\theta()$ with learnable parameters $\theta$ learns to reverse the diffusion process. 
It is also possible to model conditional distributions $p_\theta(\rvx_{0:T} \mid \rvc)$, where the output is computed as $\texttt{denoise}_\theta(\rvx_t, t, \rvc)$.
The forward process does not change in this case; in practice, we merely add the conditional signal as input to the denoising model. 

Unfortunately, we cannot train existing denoising diffusion models to sample from $p(\sig \mid \obs, \param)$, or, in fact, even from an unconditional distribution $p(\sig)$. This would require computation of the Markovian forward process in Eq.~\ref{eq:forward}. However, recall that we do not have access to any signals $\{\sig^j\}_j$ - we thus can not add any noise to any signals to then train a denoising neural network.
In other words, since no $\sig$ is directly observed, we \emph{cannot} compute  $q(\sig_t \mid \sig_{t-1})$. 

\subsection{Integrating Denoising Diffusion with Differentiable Forward Models}
We now introduce a class of denoising diffusion models that we train to directly model the distribution $p(\sig \mid \context; \paramcontext)$ over signals $\sig$ given an observation $(\context, \paramcontext)$.
Our key contribution is to directly integrate the differentiable forward model $\forwardmodel()$ into the iterative \VS{conditional} denoising process. This enables us to add noise to and denoise the observations, while nevertheless sampling the underlying signal that generates that observation.

Our model is trained on pairs of ``context'' and ``target'' observations of the same signal, denoted as $\context$ and $\target$. 
As in conventional diffusion models, for the forward process, we have $q(\target_t \mid \target_{t\text{-}1}) = \mathcal{N}(\target_t; \sqrt{1-\beta_t} \target_{t-1}, \beta_t \mathbf{I})$.
In the reverse process, we similarly denoise $\target$ conditional on $\context$:
\begin{equation}
\label{eq:target-conditioned-on-context}
    p_\theta(\mathbf{O}^\text{trgt}_{0:T} \mid \context; \paramcontext, \paramtarget) =  p(\target_T) \prod_{t=0}^T p_\theta(\target_{t-1} \mid \target_{t}, \context; \paramcontext, \paramtarget),
\end{equation}
However, unlike conventional diffusion models, we implement $p_\theta(\target_{t-1} \mid \target_{t}, \context; \paramcontext, \paramtarget)$ by first predicting an estimate of the underlying signal $\signalestimate$ and then mapping it to an estimate of the denoised observations via the differentiable $\forwardmodel$:
\begin{align}
\signalestimate &= \denoise_\theta(\context, \target_{t}; t, \paramcontext, \paramtarget),  \label{eq:z_to_x}\\
\targetestimate &= \forwardmodel(\signalestimate, \paramtarget) \label{eq:target-estimate} \\
\target_{t\text{-}1} & \sim  \mathcal{N}(\target_{t\text{-}1}; C_{t\text{-}1} \targetestimate,\; \hat{\beta}_{t\text{-}1}\mathbf{I}) \label{eq:denoise}
\end{align}
Here, $\targetestimate$ is an estimate of the \emph{clean} observation, and the constants $C_{t\text{-}1}$ and $\hat{\beta}_{t\text{-}1}$ are chosen to match the total noise added by the forward process at time $t\text{-}1$.
See Fig.~\ref{fig:graphical} for an overview. 
At test time, a signal is sampled by iterating Eq.~\ref{eq:z_to_x}, \ref{eq:target-estimate}, and \ref{eq:denoise} starting with  $p(\target_{t=T})\sim\mathcal{N}(\mathbf{0}, \mathbf{I})$.
Importantly, our models define a generative model over the underlying signal via Eq.~\ref{eq:z_to_x}: 
\begin{align}   
    \label{eq:p_signal}
    p_{\theta, \paramtarget}(\sig_{0:T} \mid \context; \paramcontext) &= \prod_{t=1}^T p_\theta(\sig_{t\text{-}1} \mid \target_{t}, \context; \paramcontext, \paramtarget).
\end{align}
\rev{We will suppress the subscript in the notation, and refer to this distribution as $p(\sig_{0:T} \mid \context; \paramcontext)$ for brevity from now.}

\textbf{Loss Function.}
\rev{We train to minimize the following two loss terms: }
\rev{
\begin{align}
    \label{eq:loss}
    \mathcal{L}_\theta^\text{trgt} =  \mathbb{E}_{\context, \target, \paramcontext, \paramtarget, t} & \Big[\| \target - \underbrace{\forwardmodel(\denoise_\theta(\context, \target_t; t, \paramcontext, \paramtarget), \paramtarget)}_{= \targetestimate} \|^2 \Big],\\
    \label{eq:novel-loss}
    \mathcal{L}_\theta^\text{novel} =  \mathbb{E}_{\context, \targetnovel, \paramcontext, \paramtarget, \paramtargetnovel, t} & \Big[\| \targetnovel - \underbrace{\forwardmodel(\denoise_\theta(\context, \target_t; t, \paramcontext, \paramtarget), \paramtargetnovel)}_{= \targetestimatenovel} \|^2 \Big].
\end{align}}
\rev{Here, we compute the estimate of the observation from the target, as well as a separate, novel forward model parameter $\paramtargetnovel$. 
In the supplemental document, we show that these losses approximate a total observation loss, maximizing the likelihood of all possible observations of the signal $\sig$.}

\textbf{Characterizing the Conditional Distribution Over Signals.} 
Due to the complexity of the reverse process, it may not be clear that the learned distribution over signals will agree with the true distribution, even in the limit of infinite data. 
However, this model will indeed asymptotically learn the true conditional distribution over signals, as we formally prove in the supplement:
\begin{proposition}
\label{prop:main-prop}
\rev{
Suppose that any signal $\sig$ can be reconstructed from the set of all \emph{all possible} observations of $\sig$.
Under this assumption, if in the limit as the number of known observations per signal goes to infinity, there are parameters $\theta$ such that $\mathcal{L}^{\text{trgt}}_\theta+\mathcal{L}^{\text{novel}}$ is minimized, then the conditional probability distribution over signals discovered by our model $p(\sig \mid \context; \paramcontext)$ agrees with the true distribution~$p^\text{true}(\sig\mid\context; \paramcontext)$.}
\end{proposition}
\rev{The proof follows by showing that our losses implicitly minimize a diffusion model loss over \emph{total observations}, which are collections of all possible observations of our signal. As such, when the observations suffice to completely reconstruct the signal, the correctness of the estimated distribution over total observations forces the estimated distribution over signals to be correct, as well.}
\begin{figure}[!t]
\centering
\includegraphics[width=\linewidth]{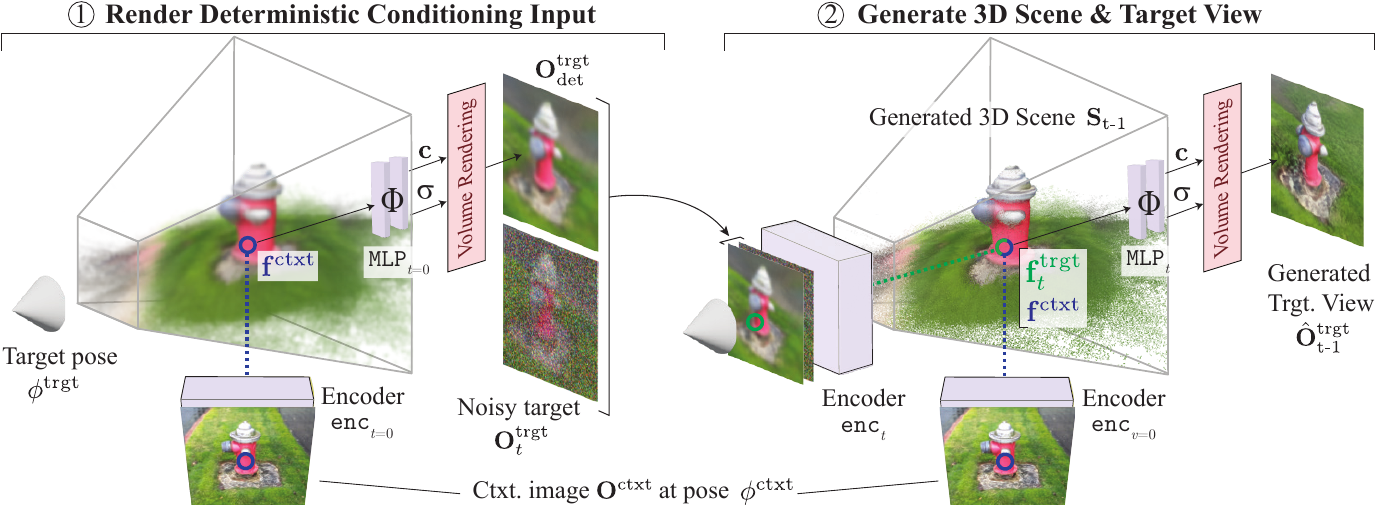}
\caption{\textbf{Overview of 3D Generative Modeling.}
We build a 3D-structured $\denoise$ operator on top of pixelNeRF~\cite{yu2021pixelnerf} that learns to sample from the distribution of 3D scenes from image observations only.
Given a context image $\context$ with camera pose $\paramcontext$, we pick a target pose $\paramtarget$. 
We render out a deterministic estimate of the depth, RGB, and features of the target view $\target_\text{det}$ using pixel-aligned features $\mathbf{f}^\text{ctxt}$ extracted from the context view with encoder $\enc_{t=0}$ (left, only RGB shown here). 
To generate a 3D scene, we concatenate the deterministic estimate with noise $\targett$, and extract features $\feat^\text{trgt}_t$ for the \emph{target} view with $\enc_{t}$. 
$\feat^\text{trgt}_t$ and $\feat^\text{ctxt}$ now jointly parameterize the radiance field of the generated scene $\signalestimate$, and we may render an estimate of the clean target view $\targetestimate$.
The model is trained end-to-end via a re-rendering loss.}
\label{fig:overview_3d}
\vspace{-0.5cm}
\end{figure} 

\section{Prior Work on Latent Variable Models for Inverse Problems}
Variational Autoencoders~\cite{kingma2013auto,rezende2014stochastic}, normalizing flows~\cite{rezende2015variational}, conditional~\cite{garnelo2018conditional} and attentive neural processes~\cite{kim2019attentive} are latent-variable models that can be combined with forward models to learn to sample from the distribution of unobserved signals from observations~\cite{kosiorek2021nerf,moreno2023laser}. However, they empirically fall short of accurately modeling complex signal distributions - in inverse graphics, for instance, such models have so far been limited to synthetic 3D scenes.
Generative Adversarial Networks can be trained with differentiable forward models in-the-loop, and have yielded impressive results in unconditional generative modeling of unobserved signals~\cite{chan2021pi,gao2022get3d,devries2021unconstrained}.
Similarly, in concurrent work, diffusion models have been leveraged for unconditional generative modeling through differentiable forward models~\cite{karnewar2023holodiffusion,mueller2022diffrf,anciukevivcius2022renderdiffusion}. 
However, unconditional models are limited to tight distributions, and no conditional generative modeling of similar quality has been demonstrated.
Diffusion models trained directly on signals have been effectively applied to diverse inverse problems such as super-resolution~\cite{chung2023diffusion, ilvr, kawar2022denoising, song2023pseudoinverse}, inpainting~\cite{song2021scorebased, chung2023diffusion, kawar2022denoising, song2023pseudoinverse}, and medical imaging~\cite{song2021solving}.
These works utilize the learned prior of the data distribution to recover the latent signal through a ``plug and play'' approach~\cite{bardsley2012mcmc, Venkatakrishnan2013PlugandPlayPF,romano2017little}, integrating the diffusion model with a forward measurement process according to Bayes' rule.
These approaches are versatile and can easily adapt to new inverse problems without retraining.
However, unlike our models, they rely on direct supervision over the signals in the form of large datasets.

\section{Applications}
We now apply our framework to three stochastic inverse problems. We focus on applications in computer vision, where we tackle the problems of inverse graphics, single-image motion prediction, and GAN inversion.
For each application, we give a detailed description of the forward model, the dataset and baselines, as well as a brief description of prior work.

\subsection{Inverse Graphics}
\begin{figure}[t]
\centering
\includegraphics[width=\linewidth]{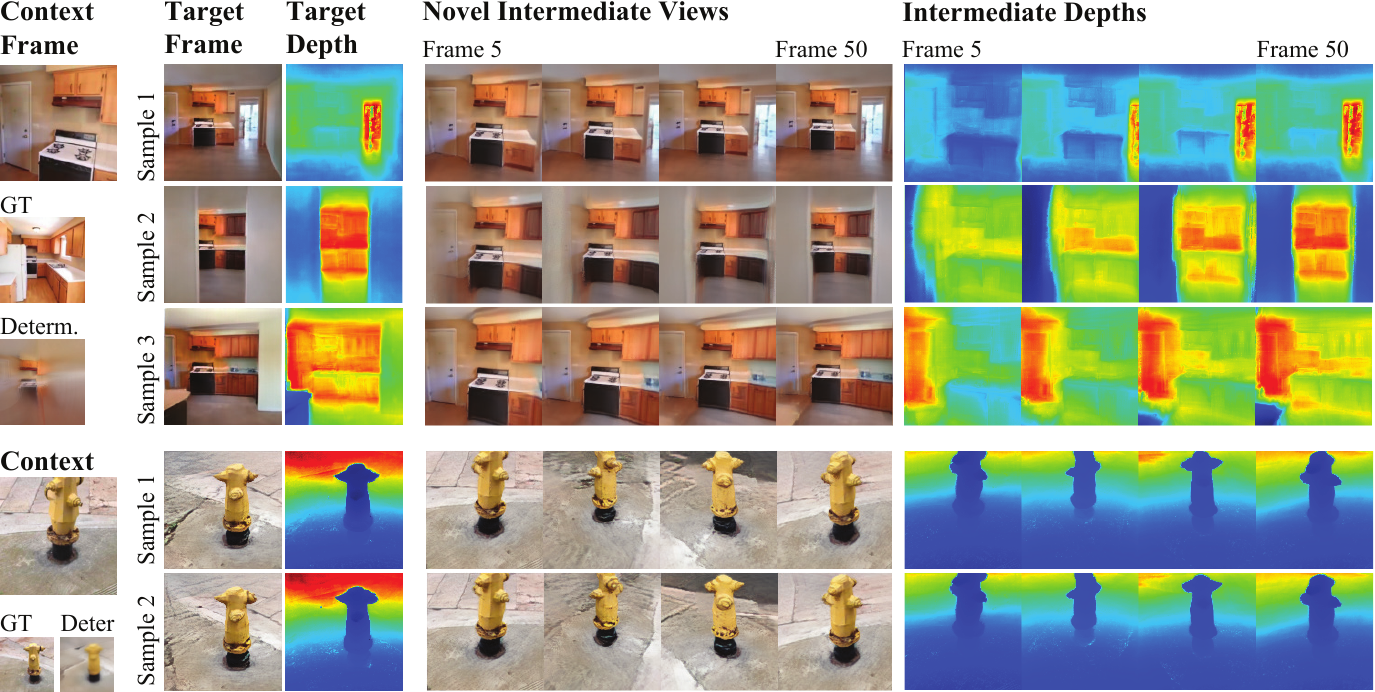}
\caption{\textbf{Sample Diversity.} We illustrate different 3D scenes sampled from the same context image for RealEstate10k and Co3D datasets. Unlike deterministic methods like pixelNeRF~\cite{yu2021pixelnerf}, our method generates diverse and distinct 3D scenes that all align with the context image.
Co3D results are generated using autoregressive sampling, where a 360 degree trajectory can be generated by iteratively sampling target images. 
Note the photorealism and diversity of the generated structures for the indoor scene, such as doors and cabinets. 
Also note the high-fidelity geometry of the occluded parts of the hydrant and the diverse background appearance. 
}
\vspace{-0.6cm}
\label{fig:3D}
\end{figure} 
We seek to learn a model that, given a single image of a 3D scene enables us to sample from the distribution over 3D scenes that are consistent with the observation. 
We expect that 3D regions visible in the image are reconstructed faithfully, while unobserved parts are generated plausibly. 
Every time we sample, we expect a \emph{different} plausible 3D generation. 
Signals $\sig$ are 3D scenes, and observations are 2D images $\obs$ and their camera parameters $\phi$. 
At training time, we assume that we have access to at least two image observations and their camera parameters per scene, such that we can assemble tuples of $(\context, \paramcontext, \target, \paramtarget)$, with 2D images $\context,\target$, and camera parameters $\paramcontext, \paramtarget$.

\textbf{Scope.} \VS{We note that our treatment of inverse graphics exceeds a mere application of the presented framework. In particular, we not only integrate the differentiable rendering $\forwardmodel$ function, but further propose a novel 3D-structured $\denoise$ function.} Here, we enable state-of-the-art conditional generation of complex, real-world 3D scenes.

\textbf{Related Work.}
Few-shot reconstruction of 3D scene representations via differentiable rendering was pioneered by deterministic methods~\cite{sitzmann2019scene,niemeyer2020dvr,henzler2021unsupervised,yu2021pixelnerf,sharma2022neural,chen2021mvsnerf,niemeyer2020dvr,du2023cross, trevithick2021grf,suhail2022generalizable,chibane2021stereo,wang2021ibrnet,lal2021coconets} that blur regions of the 3D scene unobserved in the context observations.
Probabilistic methods have been proposed that can sample from the distribution of novel views trained only on images~\cite{zhou2022sparsefusion,chan2023generative,watson2022novel,eslami2018neural,tseng2023consistent,gu2023learning}. 
While results are impressive, these methods do not allow sampling from the distribution of \emph{3D scenes}, but only from the distribution of \emph{novel views}. 
Generations are not multi-view consistent. Obtaining a 3D scene requires costly post-processing via score distillation~\cite{poole2022dreamfusion}.
Several approaches~\cite{kim2023neuralfield,mueller2022diffrf} use a two-stage design: they first reconstruct a dataset of 3D scenes, and then train a 3D diffusion model. 
However, pre-computing large 3D datasets is expensive. Further, to obtain high-quality results, dense observations are required per scene.
\VS{RenderDiffusion~\cite{anciukevivcius2022renderdiffusion} and HoloDiffusion~\cite{karnewar2023holodiffusion} integrate differentiable forward rendering with an unconditional diffusion model, enabling unconditional sampling of simple, single-object scenes.}
\AT{Similar to us, RenderDiffusion performs denoising in the image space, while HoloDiffusion uses a 3D denoising architecture.}
Other methods use priors learned by text-conditioned image diffusion models to optimize 3D scenes~\cite{melas2023realfusion,hollein2023text2room,fridman2023scenescape}.
Here, the generative model does not have explicit knowledge about the 3D information of scenes. These methods often suffer from geometric artifacts.

\textbf{Structure of $\sig$ and forward model $\render$.}
We can afford only an abridged discussion here - please see the supplement for a more detailed description.
We use NeRF~\cite{mildenhall2020nerf} as the parameterization of 3D scenes, such that $\sig$ is a function that maps a 3D coordinate $\point$ to a color $\rvc$ and density $\sigma$ as $\sig(\point) = (\sigma, \mathbf{c})$.
We require a \emph{generalizable} NeRF that is predicted in a feed-forward pass by an encoder that takes a set of $M$ context images and corresponding camera poses $\{(\obs_i, \param_i) \}_{i}^M$ as input.
We base our model on pixelNeRF~\cite{yu2021pixelnerf}.
pixelNeRF first extracts image features $\{\Feat_i\}_i$ from each context observation via an encoder $\enc$ as $\Feat_i = \enc(\obs_i)$.
Given a 3D point $\rvp$, it obtains its pixel coordinates in each context view via $\xpix_i = \pi(\rvp, \phi_i)$ via the projection operator $\pi$, and recovers a corresponding feature as $\mathbf{f}_i=\Feat_i(\xpix_i)$ by sampling the feature map at pixel coordinate $\xpix_i$.
It then parameterizes $\sig$ via an $\MLP$ as:
\begin{align}
    \sig(\rvp) = (\sigma(\point), \mathbf{c}(\point)) = \texttt{MLP}(\{(\mathbf{f}_i \oplus \point_i \}_{i}^M ),
    \label{eq:MLP}
\end{align}
where $\oplus$ is concatenation and $\point_i$ is the 3D point $\point$ transformed into the camera coordinates of observation $i$.
The number of context images $M$ is flexible, and we may condition $\sig$ on a single or several observations.
It will be convenient to refer to a pixelNeRF that is reconstructed from context and target observations $(\context, \paramcontext)$ and $(\target, \paramtarget)$ as 
\begin{align}
\sig( \cdot \mid \enc(\context), \enc(\target)),
\end{align}
where we make the pixelNeRF encoder $\enc$ explicit and drop the poses $\paramtarget$ and $\paramcontext$.
We leverage differentiable volume rendering~\cite{mildenhall2020nerf} as forward model, such that
\begin{equation}
  \obs = \render(\sig, \param),
\end{equation}
where $\sig$ is rendered from a camera with parameters $\param$. 

\begin{figure}
  \centering
  \includegraphics[width=\linewidth]{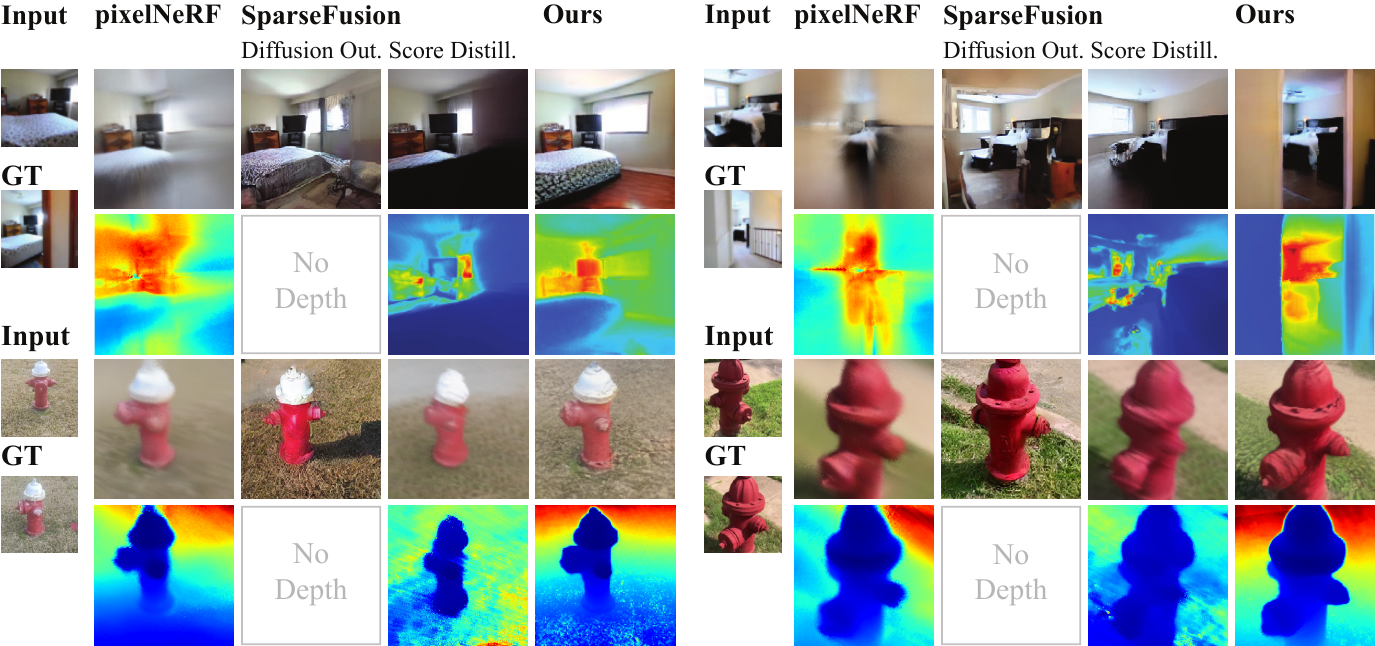}
\caption{\textbf{Qualitative Comparison for Inverse Graphics application.} We benchmark with SparseFusion~\cite{zhou2022sparsefusion} and the deterministic pixelNeRF~\cite{yu2021pixelnerf}. SparseFusion samples 2D novel views conditioned on a deterministic rendering (Diffusion Out.), and generates multi-view consistent 3D scenes only after Score Distillation. Our method consistently generates higher-quality scenes, while directly sampling 3D scenes.}
\label{fig:comparisons_co3d}
\vspace{-0.5cm}
\end{figure}

\textbf{Implementation of $\texttt{denoise}$.}
Fig.~\ref{fig:overview_3d} gives an overview of the denoising procedure.
Following our framework, we obtain the denoised target observation $\targetestimate$ as:
\begin{align}
   \label{eq:render_denoise_image}
   \targetestimate = \render(\signalestimate, \paramtarget), \quad \text{where} \\
   \signalestimate = \sig(\cdot \mid \enc_{t=0}(\context), \enct(\targett)),
   \label{eq:render_denoise}
\end{align}
where the image encoder $\enc_t$ is now conditioned on the timestep $t$.
In other words, we will generate a target view $\targetestimate$ by rendering the pixelNeRF conditioned on the context and noisy target observations.
However, feeding the noisy $\targett$ directly to pixelNeRF is insufficient. This is because the pixel-aligned features $\enc_t(\obs)$ are obtained from each view separately - thus, the features generated by $\enc_t(\targett)$ will be uninformative.
To successfully generate a 3D scene, we have to augment the $\targett$ with information from the context view.
We propose to generate conditioning information for $\targett$ by rendering a \emph{deterministic estimate}  $\targetdet = \render\left(\sig(\cdot \mid \enc_{t=0}(\context)), \paramtarget \right)$. I.e., we condition pixelNeRF only on the context view, and render an estimate of the target view via volume rendering.
However, in the extreme case of a completely uncertain target view, this results in a completely blurry image.
We thus propose to additionally render high-dimensional features. Recall that any 3D point $\point$, we have $(\sigma(\point), \mathbf{c}(\point)) = \MLP_t(\point)$. We modify $\MLP_t$ to also output a high-dimensional feature and render a deterministic feature map to augment $\targett$ (only RGB shown in figure).
We generate the final 3D scene as $\signalestimate = \sig(\cdot \mid \enc_{t=0}(\context), \enc_{t}(\targetdet \oplus \targett))$.
The final denoised target view  is then obtained according to the rendering Eq.~\ref{eq:render_denoise_image} above.

\textbf{Loss and Training.}
Our loss consists of simple least-squares terms on re-rendered views, identical to the general loss terms presented in Eqs.~\ref{eq:loss} and \ref{eq:novel-loss}, in addition to regularizers that penalize degenerate 3D scenes.
We discuss these regularizers, as well as training details, in the supplement.

\subsubsection{Results}

\textbf{Datasets}
We evaluate on two challenging real-world datasets.
We use Co3D hydrants~\cite{reizenstein21co3d} to evaluate our method on object-centric scenes. 
For scene-level 3D synthesis, we use the challenging RealEstate10k dataset~\cite{RealEstate10k}, consisting of indoor and outdoor videos of scenes. 

\textbf{Baselines}
We compare our approach with state-of-the-art approaches in deterministic and probabilistic 3D scene completion.
We use pixelNeRF as the representative method for deterministic methods that takes a single image as input and deterministically reconstructs a 3D scene. 
Our method is the first to probabilistically reconstruct 3D scenes in an end-to-end manner. 
Regardless, we compare with the concurrent SparseFusion~\cite{ding2021sparse} that learns an image-space generative model over novel views of a 3D scene. 
Score distillation of this generative model is required every time we want to obtain a multi-view consistent 3D scene, which is costly.

\textbf{Qualitative Results.}
\looseness=-1
In Fig.~\ref{fig:3D}, we show multiple samples of 3D scenes sampled from a monocular image. 
For the indoor scenes of RealEstate10k, there are large regions of uncertainty. We can sample from the distribution of valid 3D scenes, resulting in significantly different 3D scenes with plausible geometry and colors. 
The objects are faithfully reconstructed for the object-centric Co3D scenes, and the uncertainty in the scene is captured. 
We can sample larger 3D scenes and render longer trajectories by autoregressive sampling, i.e., we treat intermediate diffused images as additional context observations to sample another target observation. 
The Co3D results in Fig.~\ref{fig:3D} were generated autoregressively for a complete 360 degrees trajectory. 
In Fig.~\ref{fig:comparisons_co3d}, we compare our results with pixelNeRF~\cite{yu2021pixelnerf} and SparseFusion~\cite{zhou2022sparsefusion}. 
pixelNeRF is a deterministic method and thus leads to very blurry results in uncertain regions. 
SparseFusion reconstructs scenes by score-distillation over a 2D generative model. 
This optimization is very expensive, and does not lead to natural-looking results. 

\textbf{Quantitative Results.}
\looseness=-1
For the object-centric Co3D dataset, we evaluate the accuracy of novel views using PSNR and LPIPS~\cite{zhang2018unreasonable} metrics. 
Note that PSNR/LPIPS are not meaningful metrics for large scenes since the predictions have a large amount of uncertainty, i.e., a wide range of novel view images can be consistent with any input image. 
Thus, we report FID~\cite{heusel2017gans} and KID \cite{kid} scores to evaluate the realism of reconstructions in these cases. 
\AT{Our approach outperforms all baselines for LPIPS, FID, and KID metrics, as our model achieves more realistic results.  
We achieve slightly lower PSNR compared to pixelNeRF~\cite{yu2021pixelnerf}.
Note that PSNR favors mean estimates, and that we only evaluate our model using a single randomly sampled scene for an input image due to computational constraints. 
}

\begin{figure}[t]
    \centering
    \includegraphics[width=\textwidth]{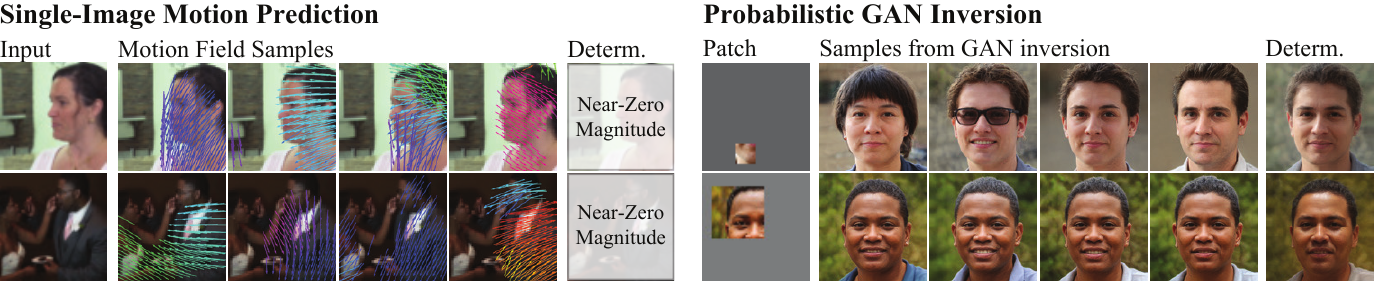}
    \caption{Qualitative Results for Single-Image Motion Prediction (left) and GAN Inversion (right).}
    \label{fig:flow_gan_results}
    \vspace{-0.5cm}
\end{figure}
\subsection{Single-Image Motion Prediction}
\begin{table}
\begin{tabular}{cc}
    \begin{minipage}[t]{.65\linewidth}
    	\vspace{0pt}
        \small
        \begin{tabular}{l|llll|ll}
                        \multicolumn{7}{l}{\hspace{-7.2pt}\small \textbf{3D Scene Completion}}\\ 
                     & \multicolumn{4}{c|}{Co3D} & \multicolumn{2}{c}{RealEstate10k}           \\ \hline %
                     & PSNR$\uparrow$    & LPIPS$\downarrow$   & FID$\downarrow$ & KID$\downarrow$  & {FID$\downarrow$} & {KID$\downarrow$}\\ \hline %
        pixelNeRF & \textbf{17.93} & 0.54 & 180.20 & 0.14 & 195.40 & 0.14 \\ 
        SparseFusion & 12.06 & 0.63 & 252.13 & 0.16 & 99.44 & 0.04 \\ 
        Ours & 17.47 & \textbf{0.42} & \textbf{84.63} & \textbf{0.05}  & \textbf{42.84} & \textbf{0.01} \\
        \end{tabular}
    \end{minipage} \hspace{0.06\linewidth}
    \begin{minipage}[t]{.25\linewidth}
    	\vspace{0pt}
        \small
        \begin{tabular}{l|ll}
                        \multicolumn{3}{l}{\hspace{-7.2pt}\small \textbf{GAN Inversion}}\\
                        & \multicolumn{2}{c}{FFHQ}\\ \hline
                     & FID$\downarrow$ & KID$\downarrow$          \\ \hline %
        Determ. &  25.7 & 0.019\\
        Ours         & \textbf{7.45} & \textbf{0.002}                               %
        \end{tabular}
    \end{minipage}
\end{tabular}
\vspace{4pt}
\caption{\textbf{Quantitative evaluation.} (left) We benchmark our 3D generative model with state-of-the-art baselines pixelNeRF~\cite{yu2021pixelnerf} and SparseFusion~\cite{zhou2022sparsefusion}. (right) We benchmark with a deterministic baseline on GAN inversion, which we drastically outperform. \label{tab:quant3D}}
\vspace{-0.8cm}
\end{table}
Here, we seek to train a model that, given a single static image, allows us to sample from \emph{all possible motions} of pixels in the image. 
Given, for instance, an image of a person performing a task, such as kicking a soccer ball, it is possible to predict potential future states. 
This is a stochastic problem, as there are multiple possible motions consistent with an image. 
We train on a dataset of natural videos~\cite{xue2019video}. We only observe RGB frames and never directly observe the underlying motion, i.e, the pixel correspondences in time are unavailable. 
We use tuples of two frames from videos within a small temporal window, and use them as our context and target observations for training.

\textbf{Related Work.}
Several papers tackle this problem, where motion in the form of optical flow~\cite{gao2018im2flow,walker2015dense,pintea2014deja}, 2D trajectories~\cite{walker2016uncertain,li2018flow}, and  human motion~\cite{walker2017pose, kanazawa2018learning} are recovered from a static image; however, all these methods assume supervision over the underlying motion.
Learning to reason about motion requires the neural network to learn about the properties and behavior of the different objects in the world. 
Thus, this serves as a useful proxy task for representation learning, and can be used as a backbone for many downstream applications~\cite{walker2016uncertain,choudhury2022guess}. 

\textbf{Structure of $\sig$ and forward model $\warp$.}
Our signal $\sig$ stores the appearance and motion information in a 2D grid. 
At any pixel $\rvu$, the signal is defined as $\sig(\rvu) = (\sig_c(\rvu), \sig_m(\rvu))$, where $\sig_c(\rvu) \in \mathbb{R}^3$ is the color value, and $\sig_m(\rvu) \in \mathbb{R}^2$ is a 2D motion vector.
The forward model is a warping operator, such that $\warp(\sig, \phi)(\rvu + \phi \sig_m(\rvu)) = \sig_c(\rvu)$ and $\phi$ is a scalar that changes the magnitude of motion.
We implement this function using a differentiable point splatting operation~\cite{Niklaus_CVPR_2020}.

\begin{wrapfigure}{r}{1.85in}
\label{fig:GAN_inversion}
	\vspace{-0.16in}
	\centering
    \includegraphics[width=1.85in]{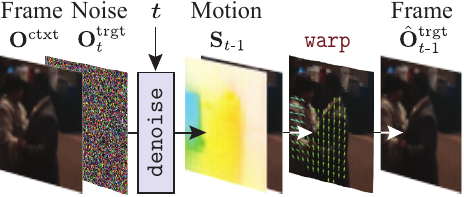}
	\vspace{-0.25in}
\end{wrapfigure}

\textbf{Implementation of $\denoise$.}
\looseness=-1
The inset figure illustrates our design. 
We use a 2D network that takes $\context$, $\target_t$, and  $t$ as input, 
and generates the motion map $\sig_m$ as the output. 
The signal is then reconstructed as $\sig = (\context, \sig_m)$.
Context and target frames correspond to parameters $\paramcontext=0$ and $\paramtarget=1$, and can be reconstructed from the signal using $\warp$.

\textbf{Loss and Evaluation.}
Similar to inverse graphics, we use reconstruction and regularization losses. 
The reconstruction losses are identical to Eqs.~\ref{eq:loss} and \ref{eq:novel-loss}, and the regularization loss is a smoothness term that encourages a natural motion of the scene, see supplement for details. 
We show results in Fig.~\ref{fig:flow_gan_results} (left), where we can estimate a diverse set of possible motion flows from monocular images. 
By smoothly interpolating $\phi$, we can generate short video sequences, even though our model only saw low-framerate video frames during training.
We also train a deterministic baseline, which only generates a single motion field. 
Due to the amount of uncertainty in this problem, the deterministic estimate collapses to a near-zero motion field regardless of the input image, and thus, fails to learn any meaningful features from images. 
\subsection{GAN Inversion}
\looseness=-1
Projecting images onto the latent space of generative adversarial networks is a well-studied problem \cite{zhu2016generative, sg2},  and enables interesting applications, as manipulating latents along known directions allows a user to effectively edit images~\cite{harkonen2020ganspace,tewari2020stylerig,shen2020interfacegan}.  
Here, we solve the problem of projecting partial images: given a small visible patch in an image, our goal is to model the distribution of possible StyleGAN2~\cite{sg2} latents that agree with the input patch. 
There are a diverse set of latents that can correspond to the input observation, and we train our method without observing supervised (\texttt{image}, \texttt{latent}) pairs. 
Instead, we train on pairs of ($\context$, $\target$) observations, where $\context$ are the small patches in images, and $\target$ are the full images.

\textbf{Related Work.} While most GAN inversion methods focus on inverting a complete image into the generator's latent space~
\cite{abdal2019image2stylegan,abdal2020image2stylegan++,bau2020semantic,
alaluf2021restyle,guan2020collaborative,pidhorskyi2020adversarial,richardson2021encoding,tov2021designing,wang2022high}, some also reconstruct GAN latents from small patches via supervised training.
Inversion is not trivial, and papers often rely on regularization~\cite{tov2021designing} or integrate the inversion with editing tasks~\cite{tewari2020pie} for higher quality.
We also integrate the inpainting task with the inversion, and seek to model the uncertainty of the GAN inversion task given only a partial observation (patch) of the target image.

\textbf{Structure of $\sig$ and forward model $\generate$.}
\looseness=-1
Our signal $\sig \in \mathbb{R}^{512}$ is a 512 dimensional latent code representing the ``$\text{w}$'' space of StyleGAN2 \cite{sg2} trained on the FFHQ \cite{sg1} dataset. 
The forward model $\generate(\sig, \phi) = \texttt{GAN}(\sig) [\phi]$ first reconstructs the image corresponding to $\sig$ using a forward pass of the GAN.
It then extracts a patch using the forward model's parameters $\phi$ that encode the patch coordinates.

\begin{wrapfigure}{r}{1.63in}
\label{fig:GAN_inversion}
	\vspace{-0.3in}
   \includegraphics[width=1.63in]{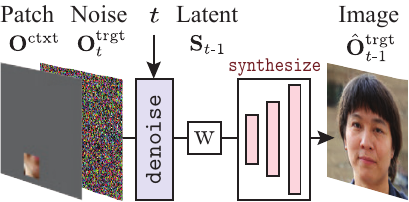}
	\vspace{-0.3in}
\end{wrapfigure}
\textbf{Implementation of $\denoise$, Loss, and Evaluation.}
Please see the inset figure for an illustration of the method.
The denoising network receives $\context$, $\target_t$, and timestep t as input, and generates an estimate of the StyleGAN latent $\mathbf{w}$. 
The loss function is identical to Eq.~\ref{eq:loss} and compares the reconstructed sample with ground truth.
We show results in Fig.~\ref{fig:flow_gan_results} (right). 
We obtain diverse samples that are all consistent with the input patch. 
We also compare with a deterministic baseline that minimizes the same loss but only produces a single estimate. 
While this deterministic estimate also agrees with the input image, it does not model the diversity of outputs. 
We consequently achieve significantly better FID \cite{heusel2017gans} and KID \cite{kid} scores than the deterministic baseline, reported in Tab.~\ref{tab:quant3D} (right).

\section{Discussion}
\textbf{Limitations.} While our method makes significant advances in generative modeling, it still has several limitations. 
Sampling 3D scenes at test time can be very slow, due to the expensive nature of the denoising process and the cost of volume rendering. 
We need multi-view observations of training scenes for the inverse graphics application. 
Our models are not trained on very large-scale datasets, and can thus not generalize to out-of-distribution data. 

\textbf{Conclusion} We have introduced a new method that tightly integrates differentiable forward models and conditional diffusion models. Our model learns to sample from the distribution of signals trained only using their observations.
We demonstrate the efficacy of our approach on three challenging computer vision problems. In inverse graphics, our method, \VS{in combination with a 3D-structured conditioning method}, enables us to directly sample from the distribution of real-world 3D scenes consistent with a single image observation. We can then render multi-view consistent novel views while obtaining diverse samples of 3D geometry and appearance in unobserved regions of the scene.
We further tackle single-image conditional motion synthesis, where we learn to sample from the distribution of 2D motion conditioned on a single image, as well as GAN inversion, where we learn to sample images that exist in the latent space of a GAN that are consistent with a given patch.
With this work, we make contributions that broaden the applicability of state-of-the-art generative modeling to a large range of scientifically relevant applications, and hope to inspire future research in this direction.

\noindent\textbf{Acknowledgements.}
This work was supported by the National Science Foundation under Grant No. 2211259, by the Singapore DSTA under DST00OECI20300823 (New Representations for Vision), by the NSF award 1955864 (Occlusion and Directional Resolution in Computational Imaging), by the ONR MURI grant N00014-22-1-2740, and by the Amazon Science Hub. We are grateful for helpful conversations with members of the Scene Representation Group David Charatan, Cameron Smith, and Boyuan Chen. We thank Zhizhuo Zhou for thoughtful discussions about the SparseFusion baseline. This article solely reflects the opinions and conclusions of its authors and no other entity.

\noindent\textbf{Author contributions.}
Ayush and Vincent conceived the idea of diffusion with forward models, designed experiments, generated most figures, and wrote most of the paper.
Ayush contributed the key insight to integrate differentiable rendering with diffusion models by denoising in image space while generating 3D scenes.
Ayush and Vincent generalized this to general forward models, and conceived the single-image motion application.
Vincent contributed the 3D-structured conditioning and generated the overview and methods figures.
Ayush wrote all initial code and ran all initial experiments. 
Ayush and Tianwei implemented the inverse graphics application and generated most of the 3D results of our model, while George helped with the baseline 3D results. Ayush executed all single-image motion experiments.
George conceived, implemented, and executed all GAN inversion experiments. %
Semon helped formalizing the method and wrote the proposition and its proof. 
Fr\'{e}do and Bill were involved in regular meetings and gave valuable feedback on results and experiments.
Josh provided intriguing cognitive science perspectives and feedback on results and experiments, and provided a significant part of the compute.
Vincent's Scene Representation Group provided a significant part of the compute, and the project profited from code infrastructure developed by and conversations with other members of the Scene Representation Group.

\clearpage
{\small
	\bibliographystyle{unsrt}
	\bibliography{bibliography}

\begin{thebibliography}{10}

\bibitem{sohl2015deep}
Jascha Sohl-Dickstein, Eric Weiss, Niru Maheswaranathan, and Surya Ganguli.
\newblock Deep unsupervised learning using nonequilibrium thermodynamics.
\newblock In {\em Proc. ICML}, 2015.

\bibitem{mueller2022diffrf}
Norman M{\"{u}}ller, , Yawar Siddiqui, Lorenzo Porzi, Samuel Rota~Bul\`{o},
  Peter Kontschieder, and Matthias Nie{\ss}ner.
\newblock Diffrf: Rendering-guided 3d radiance field diffusion.
\newblock {\em Proc. CVPR}, 2023.

\bibitem{kim2023neuralfield}
Seung~Wook Kim, Bradley Brown, Kangxue Yin, Karsten Kreis, Katja Schwarz,
  Daiqing Li, Robin Rombach, Antonio Torralba, and Sanja Fidler.
\newblock Neuralfield-ldm: Scene generation with hierarchical latent diffusion
  models.
\newblock {\em Proc. CVPR}, 2023.

\bibitem{chan2023generative}
Eric~R Chan, Koki Nagano, Matthew~A Chan, Alexander~W Bergman, Jeong~Joon Park,
  Axel Levy, Miika Aittala, Shalini De~Mello, Tero Karras, and Gordon
  Wetzstein.
\newblock Generative novel view synthesis with 3d-aware diffusion models.
\newblock {\em arXiv preprint arXiv:2304.02602}, 2023.

\bibitem{zhou2022sparsefusion}
Zhizhuo Zhou and Shubham Tulsiani.
\newblock Sparsefusion: Distilling view-conditioned diffusion for 3d
  reconstruction.
\newblock {\em Proc. CVPR}, 2023.

\bibitem{poole2022dreamfusion}
Ben Poole, Ajay Jain, Jonathan~T. Barron, and Ben Mildenhall.
\newblock Dreamfusion: Text-to-3d using 2d diffusion.
\newblock {\em Proc. ICLR}, 2023.

\bibitem{anciukevivcius2022renderdiffusion}
Titas Anciukevi{\v{c}}ius, Zexiang Xu, Matthew Fisher, Paul Henderson, Hakan
  Bilen, Niloy~J Mitra, and Paul Guerrero.
\newblock Renderdiffusion: Image diffusion for 3d reconstruction, inpainting
  and generation.
\newblock {\em Proc. CVPR}, 2023.

\bibitem{sg2}
Tero Karras, Samuli Laine, Miika Aittala, Janne Hellsten, Jaakko Lehtinen, and
  Timo Aila.
\newblock Analyzing and improving the image quality of stylegan.
\newblock In {\em Proc. CVPR}, 2020.

\bibitem{yu2021pixelnerf}
Alex Yu, Vickie Ye, Matthew Tancik, and Angjoo Kanazawa.
\newblock pixelnerf: Neural radiance fields from one or few images.
\newblock In {\em Proc. CVPR}, 2021.

\bibitem{kingma2013auto}
Diederik~P Kingma and Max Welling.
\newblock Auto-encoding variational bayes.
\newblock {\em Proc. ICLR}, 2014.

\bibitem{rezende2014stochastic}
Danilo~Jimenez Rezende, Shakir Mohamed, and Daan Wierstra.
\newblock Stochastic backpropagation and approximate inference in deep
  generative models.
\newblock In {\em Proc. ICML}, 2014.

\bibitem{rezende2015variational}
Danilo Rezende and Shakir Mohamed.
\newblock Variational inference with normalizing flows.
\newblock In {\em Proc. ICML}, 2015.

\bibitem{garnelo2018conditional}
Marta Garnelo, Dan Rosenbaum, Christopher Maddison, Tiago Ramalho, David
  Saxton, Murray Shanahan, Yee~Whye Teh, Danilo Rezende, and SM~Ali Eslami.
\newblock Conditional neural processes.
\newblock In {\em Proc. ICML}, 2018.

\bibitem{kim2019attentive}
Hyunjik Kim, Andriy Mnih, Jonathan Schwarz, Marta Garnelo, Ali Eslami, Dan
  Rosenbaum, Oriol Vinyals, and Yee~Whye Teh.
\newblock Attentive neural processes.
\newblock {\em Proc. ICLR}, 2019.

\bibitem{kosiorek2021nerf}
Adam~R Kosiorek, Heiko Strathmann, Daniel Zoran, Pol Moreno, Rosalia Schneider,
  Sona Mokr{\'a}, and Danilo~Jimenez Rezende.
\newblock Nerf-vae: A geometry aware 3d scene generative model.
\newblock In {\em Proc. ICML}, 2021.

\bibitem{moreno2023laser}
Pol Moreno, Adam~R Kosiorek, Heiko Strathmann, Daniel Zoran, Rosalia~G
  Schneider, Bj{\"o}rn Winckler, Larisa Markeeva, Th{\'e}ophane Weber, and
  Danilo~J Rezende.
\newblock Laser: Latent set representations for 3d generative modeling.
\newblock {\em arXiv preprint arXiv:2301.05747}, 2023.

\bibitem{chan2021pi}
Eric~R Chan, Marco Monteiro, Petr Kellnhofer, Jiajun Wu, and Gordon Wetzstein.
\newblock pi-gan: Periodic implicit generative adversarial networks for
  3d-aware image synthesis.
\newblock In {\em Proc. CVPR}, 2021.

\bibitem{gao2022get3d}
Jun Gao, Tianchang Shen, Zian Wang, Wenzheng Chen, Kangxue Yin, Daiqing Li,
  Or~Litany, Zan Gojcic, and Sanja Fidler.
\newblock Get3d: A generative model of high quality 3d textured shapes learned
  from images.
\newblock {\em Proc. NeurIPS}, 2022.

\bibitem{devries2021unconstrained}
Terrance DeVries, Miguel~Angel Bautista, Nitish Srivastava, Graham~W Taylor,
  and Joshua~M Susskind.
\newblock Unconstrained scene generation with locally conditioned radiance
  fields.
\newblock In {\em Proc. ICCV}, 2021.

\bibitem{karnewar2023holodiffusion}
Animesh Karnewar, Andrea Vedaldi, David Novotny, and Niloy Mitra.
\newblock Holodiffusion: Training a 3d diffusion model using 2d images.
\newblock {\em Proc. CVPR}, 2023.

\bibitem{chung2023diffusion}
Hyungjin Chung, Jeongsol Kim, Michael~Thompson Mccann, Marc~Louis Klasky, and
  Jong~Chul Ye.
\newblock Diffusion posterior sampling for general noisy inverse problems.
\newblock In {\em Proc. ICLR}, 2023.

\bibitem{ilvr}
Jooyoung Choi, Sungwon Kim, Yonghyun Jeong, Youngjune Gwon, and Sungroh Yoon.
\newblock Ilvr: Conditioning method for denoising diffusion probabilistic
  models.
\newblock {\em Proc. ICCV}, 2021.

\bibitem{kawar2022denoising}
Bahjat Kawar, Michael Elad, Stefano Ermon, and Jiaming Song.
\newblock Denoising diffusion restoration models.
\newblock In {\em Proc. NeurIPS}, 2022.

\bibitem{song2023pseudoinverse}
Jiaming Song, Arash Vahdat, Morteza Mardani, and Jan Kautz.
\newblock Pseudoinverse-guided diffusion models for inverse problems.
\newblock In {\em Proc. ICLR}, 2023.

\bibitem{song2021scorebased}
Yang Song, Jascha Sohl-Dickstein, Diederik~P Kingma, Abhishek Kumar, Stefano
  Ermon, and Ben Poole.
\newblock Score-based generative modeling through stochastic differential
  equations.
\newblock In {\em Proc. ICLR}, 2021.

\bibitem{song2021solving}
Yang Song, Liyue Shen, Lei Xing, and Stefano Ermon.
\newblock Solving inverse problems in medical imaging with score-based
  generative models.
\newblock {\em Proc. ICLR}, 2022.

\bibitem{bardsley2012mcmc}
Johnathan~M Bardsley.
\newblock Mcmc-based image reconstruction with uncertainty quantification.
\newblock {\em SIAM Journal on Scientific Computing}, 34(3):A1316--A1332, 2012.

\bibitem{Venkatakrishnan2013PlugandPlayPF}
Singanallur Venkatakrishnan, Charles~A. Bouman, and Brendt Wohlberg.
\newblock Plug-and-play priors for model based reconstruction.
\newblock {\em 2013 IEEE Global Conference on Signal and Information
  Processing}, pages 945--948, 2013.

\bibitem{romano2017little}
Yaniv Romano, Michael Elad, and Peyman Milanfar.
\newblock The little engine that could: Regularization by denoising (red).
\newblock {\em SIAM Journal on Imaging Sciences}, 10(4):1804--1844, 2017.

\bibitem{sitzmann2019scene}
Vincent Sitzmann, Michael Zollh{\"o}fer, and Gordon Wetzstein.
\newblock Scene representation networks: Continuous 3d-structure-aware neural
  scene representations.
\newblock {\em Proc. NeurIPS}, 2019.

\bibitem{niemeyer2020dvr}
Michael Niemeyer, Lars Mescheder, Michael Oechsle, and Andreas Geiger.
\newblock Differentiable volumetric rendering: Learning implicit 3d
  representations without 3d supervision.
\newblock In {\em Proc. CVPR}, 2020.

\bibitem{henzler2021unsupervised}
Philipp Henzler, Jeremy Reizenstein, Patrick Labatut, Roman Shapovalov, Tobias
  Ritschel, Andrea Vedaldi, and David Novotny.
\newblock Unsupervised learning of 3d object categories from videos in the
  wild.
\newblock In {\em Proc. CVPR}, 2021.

\bibitem{sharma2022neural}
Prafull Sharma, Ayush Tewari, Yilun Du, Sergey Zakharov, Rares~Andrei Ambrus,
  Adrien Gaidon, William~T Freeman, Fredo Durand, Joshua~B Tenenbaum, and
  Vincent Sitzmann.
\newblock Neural groundplans: Persistent neural scene representations from a
  single image.
\newblock In {\em Proc. ICLR}.

\bibitem{chen2021mvsnerf}
Anpei Chen, Zexiang Xu, Fuqiang Zhao, Xiaoshuai Zhang, Fanbo Xiang, Jingyi Yu,
  and Hao Su.
\newblock Mvsnerf: Fast generalizable radiance field reconstruction from
  multi-view stereo.
\newblock In {\em Proc. ICCV}, 2021.

\bibitem{du2023cross}
Yilun Du, Cameron Smith, Ayush Tewari, and Vincent Sitzmann.
\newblock Learning to render novel views from wide-baseline stereo pairs.
\newblock In {\em Proc. CVPR}, 2023.

\bibitem{trevithick2021grf}
Alex Trevithick and Bo~Yang.
\newblock Grf: Learning a general radiance field for 3d representation and
  rendering.
\newblock In {\em Proc. ICCV}, 2021.

\bibitem{suhail2022generalizable}
Mohammed Suhail, Carlos Esteves, Leonid Sigal, and Ameesh Makadia.
\newblock Generalizable patch-based neural rendering.
\newblock In {\em Proc. ECCV}, 2022.

\bibitem{chibane2021stereo}
Julian Chibane, Aayush Bansal, Verica Lazova, and Gerard Pons-Moll.
\newblock Stereo radiance fields (srf): Learning view synthesis for sparse
  views of novel scenes.
\newblock In {\em Proc. CVPR}, 2021.

\bibitem{wang2021ibrnet}
Qianqian Wang, Zhicheng Wang, Kyle Genova, Pratul~P Srinivasan, Howard Zhou,
  Jonathan~T Barron, Ricardo Martin-Brualla, Noah Snavely, and Thomas
  Funkhouser.
\newblock Ibrnet: Learning multi-view image-based rendering.
\newblock In {\em Proc. CVPR}, 2021.

\bibitem{lal2021coconets}
Shamit Lal, Mihir Prabhudesai, Ishita Mediratta, Adam~W Harley, and Katerina
  Fragkiadaki.
\newblock Coconets: Continuous contrastive 3d scene representations.
\newblock In {\em Proc. CVPR}, 2021.

\bibitem{watson2022novel}
Daniel Watson, William Chan, Ricardo Martin-Brualla, Jonathan Ho, Andrea
  Tagliasacchi, and Mohammad Norouzi.
\newblock Novel view synthesis with diffusion models.
\newblock {\em Proc. ICLR}, 2023.

\bibitem{eslami2018neural}
SM~Ali Eslami, Danilo Jimenez~Rezende, Frederic Besse, Fabio Viola, Ari~S
  Morcos, Marta Garnelo, Avraham Ruderman, Andrei~A Rusu, Ivo Danihelka, Karol
  Gregor, et~al.
\newblock Neural scene representation and rendering.
\newblock {\em Science}, 360(6394):1204--1210, 2018.

\bibitem{tseng2023consistent}
Hung-Yu Tseng, Qinbo Li, Changil Kim, Suhib Alsisan, Jia-Bin Huang, and
  Johannes Kopf.
\newblock Consistent view synthesis with pose-guided diffusion models.
\newblock {\em arXiv preprint arXiv:2303.17598}, 2023.

\bibitem{gu2023learning}
Jiatao Gu, Qingzhe Gao, Shuangfei Zhai, Baoquan Chen, Lingjie Liu, and Josh
  Susskind.
\newblock Learning controllable 3d diffusion models from single-view images.
\newblock {\em arXiv preprint arXiv:2304.06700}, 2023.

\bibitem{melas2023realfusion}
Luke Melas-Kyriazi, Christian Rupprecht, Iro Laina, and Andrea Vedaldi.
\newblock Realfusion: 360 $\{$$\backslash$deg$\}$ reconstruction of any object
  from a single image.
\newblock {\em Proc. CVPR}, 2023.

\bibitem{hollein2023text2room}
Lukas H{\"o}llein, Ang Cao, Andrew Owens, Justin Johnson, and Matthias
  Nie{\ss}ner.
\newblock Text2room: Extracting textured 3d meshes from 2d text-to-image
  models.
\newblock {\em arXiv preprint arXiv:2303.11989}, 2023.

\bibitem{fridman2023scenescape}
Rafail Fridman, Amit Abecasis, Yoni Kasten, and Tali Dekel.
\newblock Scenescape: Text-driven consistent scene generation.
\newblock {\em arXiv preprint arXiv:2302.01133}, 2023.

\bibitem{mildenhall2020nerf}
Ben Mildenhall, Pratul~P. Srinivasan, Matthew Tancik, Jonathan~T. Barron, Ravi
  Ramamoorthi, and Ren Ng.
\newblock Nerf: Representing scenes as neural radiance fields for view
  synthesis.
\newblock In {\em Proc. ECCV}, 2020.

\bibitem{reizenstein21co3d}
Jeremy Reizenstein, Roman Shapovalov, Philipp Henzler, Luca Sbordone, Patrick
  Labatut, and David Novotny.
\newblock Common objects in 3d: Large-scale learning and evaluation of
  real-life 3d category reconstruction.
\newblock In {\em Proc. ICCV}, 2021.

\bibitem{RealEstate10k}
Tinghui Zhou, Richard Tucker, John Flynn, Graham Fyffe, and Noah Snavely.
\newblock Stereo magnification: Learning view synthesis using multiplane
  images.
\newblock {\em ACM Trans. Graph. (Proc. SIGGRAPH)}, 37, 2018.

\bibitem{ding2021sparse}
Yi~Ding, Alex Rich, Mason Wang, Noah Stier, Matthew Turk, Pradeep Sen, and
  Tobias H{\"o}llerer.
\newblock Sparse fusion for multimodal transformers.
\newblock {\em arXiv preprint arXiv:2111.11992}, 2021.

\bibitem{zhang2018unreasonable}
Richard Zhang, Phillip Isola, Alexei~A Efros, Eli Shechtman, and Oliver Wang.
\newblock The unreasonable effectiveness of deep features as a perceptual
  metric.
\newblock In {\em Proc. CVPR}, 2018.

\bibitem{heusel2017gans}
Martin Heusel, Hubert Ramsauer, Thomas Unterthiner, Bernhard Nessler, and Sepp
  Hochreiter.
\newblock Gans trained by a two time-scale update rule converge to a local nash
  equilibrium.
\newblock {\em Proc. NeurIPS}, 2017.

\bibitem{kid}
Miko{\l}aj Bi{\'n}kowski, Danica~J Sutherland, Michael Arbel, and Arthur
  Gretton.
\newblock Demystifying mmd gans.
\newblock {\em arXiv preprint arXiv:1801.01401}, 2018.

\bibitem{xue2019video}
Tianfan Xue, Baian Chen, Jiajun Wu, Donglai Wei, and William~T Freeman.
\newblock Video enhancement with task-oriented flow.
\newblock {\em International Journal of Computer Vision}, 127:1106--1125, 2019.

\bibitem{gao2018im2flow}
Ruohan Gao, Bo~Xiong, and Kristen Grauman.
\newblock Im2flow: Motion hallucination from static images for action
  recognition.
\newblock In {\em Proc. CVPR}, 2018.

\bibitem{walker2015dense}
Jacob Walker, Abhinav Gupta, and Martial Hebert.
\newblock Dense optical flow prediction from a static image.
\newblock In {\em Proc. ICCV}, 2015.

\bibitem{pintea2014deja}
Silvia~L Pintea, Jan~C van Gemert, and Arnold~WM Smeulders.
\newblock D{\'e}ja vu: Motion prediction in static images.
\newblock In {\em Proc. ECCV}, 2014.

\bibitem{walker2016uncertain}
Jacob Walker, Carl Doersch, Abhinav Gupta, and Martial Hebert.
\newblock An uncertain future: Forecasting from static images using variational
  autoencoders.
\newblock In {\em Proc. ECCV}. Springer, 2016.

\bibitem{li2018flow}
Yijun Li, Chen Fang, Jimei Yang, Zhaowen Wang, Xin Lu, and Ming-Hsuan Yang.
\newblock Flow-grounded spatial-temporal video prediction from still images.
\newblock In {\em Proc. ICCV}, 2018.

\bibitem{walker2017pose}
Jacob Walker, Kenneth Marino, Abhinav Gupta, and Martial Hebert.
\newblock The pose knows: Video forecasting by generating pose futures.
\newblock In {\em Proc. ICCV}, 2017.

\bibitem{kanazawa2018learning}
Angjoo Kanazawa, Shubham Tulsiani, Alexei~A Efros, and Jitendra Malik.
\newblock Learning category-specific mesh reconstruction from image
  collections.
\newblock In {\em Proc. ECCV}, 2018.

\bibitem{choudhury2022guess}
Subhabrata Choudhury, Laurynas Karazija, Iro Laina, Andrea Vedaldi, and
  Christian Rupprecht.
\newblock Guess what moves: unsupervised video and image segmentation by
  anticipating motion.
\newblock {\em arXiv preprint arXiv:2205.07844}, 2022.

\bibitem{Niklaus_CVPR_2020}
Simon Niklaus and Feng Liu.
\newblock Softmax splatting for video frame interpolation.
\newblock In {\em Proc. CVPR}, 2020.

\bibitem{zhu2016generative}
Jun-Yan Zhu, Philipp Kr{\"a}henb{\"u}hl, Eli Shechtman, and Alexei~A. Efros.
\newblock Generative visual manipulation on the natural image manifold.
\newblock In {\em Proc. ECCV}, 2016.

\bibitem{harkonen2020ganspace}
Erik H{\"a}rk{\"o}nen, Aaron Hertzmann, Jaakko Lehtinen, and Sylvain Paris.
\newblock Ganspace: Discovering interpretable gan controls.
\newblock {\em Proc. NeurIPS}, 2020.

\bibitem{tewari2020stylerig}
Ayush Tewari, Mohamed Elgharib, Gaurav Bharaj, Florian Bernard, Hans-Peter
  Seidel, Patrick P{\'e}rez, Michael Zollhofer, and Christian Theobalt.
\newblock Stylerig: Rigging stylegan for 3d control over portrait images.
\newblock In {\em Proc. CVPR}, 2020.

\bibitem{shen2020interfacegan}
Yujun Shen, Ceyuan Yang, Xiaoou Tang, and Bolei Zhou.
\newblock Interfacegan: Interpreting the disentangled face representation
  learned by gans.
\newblock {\em IEEE transactions on pattern analysis and machine intelligence},
  44(4):2004--2018, 2020.

\bibitem{abdal2019image2stylegan}
Rameen Abdal, Yipeng Qin, and Peter Wonka.
\newblock Image2stylegan: How to embed images into the stylegan latent space?
\newblock In {\em Proc. ICCV}, 2019.

\bibitem{abdal2020image2stylegan++}
Rameen Abdal, Yipeng Qin, and Peter Wonka.
\newblock Image2stylegan++: How to edit the embedded images?
\newblock In {\em Proc. CVPR}, 2020.

\bibitem{bau2020semantic}
David Bau, Hendrik Strobelt, William Peebles, Jonas Wulff, Bolei Zhou, Jun-Yan
  Zhu, and Antonio Torralba.
\newblock Semantic photo manipulation with a generative image prior.
\newblock {\em arXiv preprint arXiv:2005.07727}, 2020.

\bibitem{alaluf2021restyle}
Yuval Alaluf, Or~Patashnik, and Daniel Cohen-Or.
\newblock Restyle: A residual-based stylegan encoder via iterative refinement.
\newblock In {\em Proc. ICCV}, 2021.

\bibitem{guan2020collaborative}
Shanyan Guan, Ying Tai, Bingbing Ni, Feida Zhu, Feiyue Huang, and Xiaokang
  Yang.
\newblock Collaborative learning for faster stylegan embedding.
\newblock {\em arXiv preprint arXiv:2007.01758}, 2020.

\bibitem{pidhorskyi2020adversarial}
Stanislav Pidhorskyi, Donald~A Adjeroh, and Gianfranco Doretto.
\newblock Adversarial latent autoencoders.
\newblock In {\em Proc. CVPR}, 2020.

\bibitem{richardson2021encoding}
Elad Richardson, Yuval Alaluf, Or~Patashnik, Yotam Nitzan, Yaniv Azar, Stav
  Shapiro, and Daniel Cohen-Or.
\newblock Encoding in style: a stylegan encoder for image-to-image translation.
\newblock In {\em Proc. CVPR}, 2021.

\bibitem{tov2021designing}
Omer Tov, Yuval Alaluf, Yotam Nitzan, Or~Patashnik, and Daniel Cohen-Or.
\newblock Designing an encoder for stylegan image manipulation.
\newblock {\em ACM Transactions on Graphics (TOG)}, 40(4):1--14, 2021.

\bibitem{wang2022high}
Tengfei Wang, Yong Zhang, Yanbo Fan, Jue Wang, and Qifeng Chen.
\newblock High-fidelity gan inversion for image attribute editing.
\newblock In {\em Proc. CVPR}, 2022.

\bibitem{tewari2020pie}
Ayush Tewari, Mohamed Elgharib, Florian Bernard, Hans-Peter Seidel, Patrick
  P{\'e}rez, Michael Zollh{\"o}fer, and Christian Theobalt.
\newblock Pie: Portrait image embedding for semantic control.
\newblock {\em ACM Transactions on Graphics (TOG)}, 39(6):1--14, 2020.

\bibitem{sg1}
Tero Karras, Samuli Laine, and Timo Aila.
\newblock A style-based generator architecture for generative adversarial
  networks.
\newblock In {\em Proc. CVPR}, 2019.

\end{thebibliography}
}



\section*{Supplementary material}

# Diffusion with Forward Models: Solving Stochastic Inverse Problems Without Direct Supervision

Ayush Tewari[1*]  Tianwei Yin[1*]  George Cazenavette[1]  Semon Rezchikov[4]
Joshua B. Tenenbaum[1,2,3]  Frédo Durand[1]  William T. Freeman[1]  Vincent Sitzmann[1]

[1]MIT CSAIL  [2]MIT BCS  [3]MIT CBMM  [4]Princeton IAS

## Contents

## 1 Proposition

**Proposition 1.** Suppose that any signal $\mathbf{S}$ can be reconstructed from the set of all *all possible* observations of $\mathbf{S}$. Under this assumption, if in the limit as the number of known observations per signal goes to infinity, there are parameters $\theta$ such that $\mathcal{L}_\theta^{\text{trgt}} + \mathcal{L}^{\text{novel}}$ is minimized, then the conditional probability distribution over signals discovered by our model $p(\mathbf{S} \mid \mathbf{O}^{\text{ctxt}}; \phi^{\text{ctxt}})$ agrees with the true distribution $p^{\text{true}}(\mathbf{S} \mid \mathbf{O}^{\text{ctxt}}; \phi^{\text{ctxt}})$.

The total observation loss is defined in Equation equation 4 below.

After introducing some notation, we will formalize the assumptions made in the proposition.

**Definition 1.** We call the collection of all observations that correspond to a signal a *total observation* of the signal $\mathbf{O}^{\text{total}}$. Formally,

$$\mathbf{O}^{\text{total}} = \mathbf{O}^{\text{total}}(\mathbf{S}) = \{(\phi, \texttt{forward}(\mathbf{S}, \phi))\}_{\phi \in \mathcal{P}}.$$

Here, $\mathcal{P}$ denotes the set of *parameters* of the forward model, e.g. $\mathcal{P} = SE(3)$ for the inverse graphics application in the paper.

**Definition 2.** We define the *scattering map* as the (measurable) map sending signal $\mathbf{S}$ to its total image $\mathbf{O}^{\text{total}}$:

$$Scatter : \mathbf{S} \mapsto \mathbf{O}^{\text{total}}(\mathbf{S}).$$

For a reference for the technical notion of a measurable map, see any textbook on measure theory (e.g. [1]); all maps arising in machine-learning models are measurable because they are piecewise continuous.

---

*Equal Contribution.

**Assumption**  We formalize the assumption of Proposition 1 by requiring that there is a (measurable) map $Scatter^{-1}$ from total observations to signals which satisfies, for all signals under consideration,

$$Scatter^{-1}(Scatter(\mathbf{S})) = \mathbf{S}. \quad (1)$$

In other words, given all possible observations of a signal, we can uniquely reconstruct the signal (for the class of signals under consideration). Alternatively, the map $Scatter$ is injective. This assumption is a basic assumption necessary for many algorithms in 3D computer vision, and underlies the recent success of differentiable rendering for 3D scene reconstruction [2] from large sets of image observations. Note that there may be total observations $\mathbf{O}^{\text{total}} = \{(\phi, \mathbf{O}^{\text{total}}_\phi)\}_{\phi \in \mathcal{P}}$ which do *not* arise as the total observations $\mathbf{O}^{\text{total}}(S)$ of any signal $\mathbf{S}$. Equation 1 makes no assumption on the behavior of $Scatter^{-1}(\mathbf{O}^{\text{total}})$ on such 'inconsistent' total observations $\mathbf{O}^{\text{total}}$.

**Observations generated by our model are slices of total observations.**  A basic property of our model is that the target observations arise from predicted signals, since

$$\mathbf{O}^{\text{trgt}} = \texttt{forward}(\mathbf{S}, \phi^{\text{trgt}}).$$

Thus, our model is limited to modeling the space over observations that are a member of the total observations set, i.e., $(\phi^{\text{trgt}}, \mathbf{O}^{\text{trgt}}) \in \mathbf{O}^{\text{total}}(\mathbf{S})$ for some signal $\mathbf{S}$. This is an important property that is not trivially true for many existing models, e.g., for inverse graphics, many light-field-based approaches [3–6] do not satisfy this property.

**The predicted distribution over signals can be recovered from the distribution over observations.**
Since we can reconstruct a signal from its total observation, we have that $Scatter^{-1}(Scatter(U)) = U$ for any set of signals $U$. Writing

$$V = \{(\phi^{\text{trgt}}, \texttt{forward}(\mathbf{S}, \phi^{\text{trgt}})) \mid \phi^{\text{trgt}} \in \mathcal{P}, \mathbf{S} \in U\} = \{Scatter(\mathbf{S}) \mid \mathbf{S} \in U\}.$$

for the set of *total observations* of signals $\mathbf{S} \in U$, we therefore have that

$$p(\mathbf{O}^{\text{total}}(\mathbf{S}) \in V \mid \mathbf{O}^{\text{ctxt}}; \phi^{\text{ctxt}}) = p(\mathbf{S} \in U \mid \mathbf{O}^{\text{ctxt}}; \phi^{\text{ctxt}}). \quad (2)$$

As such, we can recover $p(\mathbf{S} \in U \mid \mathbf{O}^{\text{ctxt}}; \phi^{\text{ctxt}})$ by computing $p(\mathbf{O}^{\text{total}}(\mathbf{S}) \in V \mid \mathbf{O}^{\text{ctxt}}; \phi^{\text{ctxt}})$ for all possible $V$ (where we note that if $V$ consists of total observations that do not arise from signals then its probability is zero).

**Our loss maximizes the likelihood over total observations.**  We now claim that the loss we optimize forces our model to find parameters $\theta$ such that

$$p(\mathbf{O}^{\text{total}}(\mathbf{S}) \in V \mid \mathbf{O}^{\text{ctxt}}; \phi^{\text{ctxt}}) = p^{true}(\mathbf{O}^{\text{total}}(\mathbf{S}) \in V \mid \mathbf{O}^{\text{ctxt}}; \phi^{\text{ctxt}}) \quad (3)$$

We first define the *total observation loss*

$$\mathcal{L}^{\text{total}}_\theta = \mathbb{E}_{\mathbf{O}^{\text{ctxt}}, \mathbf{O}^{\text{trgt}}, \phi^{\text{ctxt}}, \phi^{\text{trgt}}, t} \left[ \|\mathbf{O}^{\text{trgt}} - \texttt{forward}(\texttt{denoise}_\theta(\mathbf{O}^{\text{ctxt}}, \mathbf{O}^{\text{total}}_t; t, \phi^{\text{ctxt}}), \phi^{\text{trgt}})\|^2 \right] \quad (4)$$

This is the same as $\mathcal{L}^{\text{target}}_\theta$ of the main text, but with $\texttt{denoise}_\theta$ depending on the total observation. We now have the identity

$$\mathbb{E}_{\mathbf{O}^{\text{trgt}}, \phi^{\text{trgt}}} \left[ \|\mathbf{O}^{\text{trgt}} - \texttt{forward}(\texttt{denoise}_\theta(\mathbf{O}^{\text{ctxt}}, \mathbf{O}^{\text{total}}_t; t, \phi^{\text{ctxt}}), \phi^{\text{trgt}})\|^2 \right]$$
$$= \|\mathbf{O}^{\text{total}}_t - \hat{\mathbf{O}}^{\text{total}}_{t-1}\|^2 \quad (5)$$
$$= C_t D_{KL}(q(\mathbf{O}^{\text{total}}_{t-1} \mid \mathbf{O}^{\text{total}}_t, \mathbf{O}^{\text{total}}_0, \mathbf{O}^{\text{ctxt}}; \phi^{\text{trgt}}) \mid p_\theta(\mathbf{O}^{\text{total}}_{t-1} \mid \mathbf{O}^{\text{total}}_t, \mathbf{O}^{\text{ctxt}}; \phi^{\text{ctxt}}).$$

where $C_t$ is some positive constant for each $t$; this follows from Equations 95-99 of [7]. Thus, *if the model has parameters $\theta$ such that Eq. 3 holds for this parameter, then this will also hold the global minimum of the loss,* since Eq. 3 holds exactly when the $t = 1$ term of Eq. 5 is zero.

It is natural to train such a model by minimizing the loss

$$\|\mathbf{O}^{\text{trgt}} - \underbrace{\texttt{forward}(\texttt{denoise}_\theta(\mathbf{O}^{\text{ctxt}}, \mathbf{O}^{\text{total}}_t; t, \phi^{\text{ctxt}}), \phi^{\text{trgt}})}_{=\hat{\mathbf{O}}^{\text{trgt}}_{t\text{-}1}}\|^2$$

with randomly-sampled forward-model parameters $\phi^{\text{trgt}}$. This is what we do in our real training procedure, except that $\texttt{denoise}_\theta$ now only depends on a slice of $\mathbf{O}_t^{\text{total}}$, namely $\mathbf{O}_t^{\text{trgt}}$, as well as on $\phi^{\text{trgt}}$, see Eq. 8 in the main paper. Now, nothing in equation 5 requires $\hat{\mathbf{O}}_{t-1}^{\text{total}}$ to depend on all of $\mathbf{O}_t^{\text{total}}$; the equation is still valid even if $\hat{\mathbf{O}}_{t-1}^{\text{total}}$ is a function only of a slice of $\mathbf{O}_t^{\text{total}}$. This is precisely the case in our training procedure. As such, the addition of the term $\mathcal{L}_{\text{novel}}$ to the loss (see Eq. 9 in the main paper), when $\phi^{\text{novel}}$ is stochastically sampled, *forces the quantity in Equation 5 to be minimized* even though our estimate of the denoised signal only depends on the context observation and the noised target observation. Thus, the conclusion of this proposition still applies to our training procedure.

**Conclusion of proof.**  We now conclude that
$$p(\mathbf{S} \in U \mid \mathbf{O}^{\text{ctxt}}; \phi^{\text{ctxt}}) = p^{true}(\mathbf{S} \in U \mid \mathbf{O}^{\text{ctxt}}; \phi^{\text{ctxt}}) \qquad (6)$$
by applying (3) and using the fact that (2) holds both for $p$ and for $p^{true}$. Equation 6 is the desired conclusion.

**Remark on 3D consistency.**  Technically, in the models in our paper, we have that $p(\mathbf{S} \mid \mathbf{O}^{\text{ctxt}}; \phi^{\text{ctxt}})$ is actually dependent on auxiliary parameter $\phi^{\text{trgt}}$: we make predictions over signals using a diffusion model coupled with a *particular choice* of forward-model parameter. As such, to be precise, we say that our model predicts a family of distributions $p_{\phi^{\text{trgt}}}(\mathbf{S} \mid \mathbf{O}^{\text{ctxt}}; \phi^{\text{ctxt}})$ depending on $\phi^{\text{trgt}}$, where these distributions may differ for different values of $\phi^{\text{trgt}}$. Correspondingly, the model predicts a family of distributions $p_{\phi^{\text{trgt}}}(\mathbf{O}^{\text{total}} \mid \mathbf{O}^{\text{ctxt}}; \phi^{\text{trgt}})$. However, the addition of the $\mathcal{L}_{\text{novel}}$ term forces learned distribution over signals to agree with the true distribution over signals in the limit of infinite observations; as such, in that same limit, the learned distribution over signals becomes independent of $\phi^{\text{trgt}}$ since the true distribution is manifestly independent of it.

**Inverting the scatter map is unnecessary.**  In the above argument, while we assumed the inverse to the $Scatter$ map, we did not need to *compute* the map $Scatter^{-1}$ to argue that the estimated probability densities $p(\mathbf{S} \mid \mathbf{O}^{\text{ctxt}}; \phi^{\text{ctxt}})$ agree with $p^{true}(\mathbf{S} \mid \mathbf{O}^{\text{ctxt}}; \phi^{\text{ctxt}})$. This is a highly desirable property, as the map $Scatter^{-1}$ often cannot be computed efficiently. Thus, our model learns correct estimates of $p^{true}(\mathbf{S} \mid \mathbf{O}^{\text{ctxt}}; \phi^{\text{ctxt}})$ without ever explicitly computing $Scatter^{-1}$.

## 2 Details of the Method

### 2.1 Inverse Graphics

**Loss Function**  We incorporate the use of several regularization terms:
$$\mathcal{L}_{\text{reg}} = \mathcal{L}_{\text{LPIPS}} + \mathcal{L}_{\text{depth}} + \mathcal{L}_{\text{cond}}, \qquad (7)$$
$$\mathcal{L}_{\text{LPIPS}} = \mathcal{L}_{\text{LPIPS}}(\hat{\mathbf{O}}_{t\text{-}1}^{\text{trgt}}, \mathbf{O}^{\text{trgt}}), \qquad (8)$$
$$\mathcal{L}_{\text{depth}} = \mathcal{L}_{\text{EAS}} + \mathcal{L}_{\text{dist}}, \qquad (9)$$
$$\mathcal{L}_{\text{cond}} = \|\mathbf{O}_{\text{det}}^{\text{trgtcolor}} - \mathbf{O}^{\text{trgt}}\|^2. \qquad (10)$$

Here, $\mathcal{L}_{\text{LPIPS}}$ is the LPIPS perceptual loss [8] that encourages rendered images to be perceptually similar to the ground truth observation. This has been shown to help improve the quality of diffusion models [9]. We further regularize the depth renderings from the target and novel viewpoints using an edge-aware smoothness loss [10] and a distortion loss [11] that discourages floating geometry artifacts. These depth regularization terms encourage *natural* 3D geometry reconstructions. Finally, we use $\mathcal{L}_{\text{cond}}$ on the rgb component of the deterministic estimate $\mathbf{O}_{\text{det}}^{\text{trgt}}$, denoted as $\mathbf{O}_{\text{det}}^{\text{trgtcolor}}$. Recall that we use $\mathbf{O}_{\text{det}}^{\text{trgt}}$ to condition our denoising network, and that it includes color as well as high-dimensional features. We use multiplier hyperparameters 0.2 for $\mathcal{L}_{\text{LPIPS}}$ and 0.02 for $\mathcal{L}_{\text{depth}}$. Our code will be publicly released to aid in reproducibility.

**Remark on regularization**  Recall our assumption in Proposition 1 that the map from all observations to the signal is invertible. In the 3D setting, where we use real-world 2D datasets for training, we do not have access to all possible signal observations. On such training data, this assumption is not strictly true, since multiple 3D scenes can explain a subset of observations. However, the addition of the regularizing terms singles out *preferred* choices of 3D scenes explaining the known observations.

17

Heuristically, our depth and smoothness regularizers $\mathcal{L}_{\text{depth}}$ make the map between scenes and the observations in the training dataset invertible, i.e., they single out a *uniquely determined natural* 3D scene that explains the observations in the impoverished 2D dataset.

**Training Details** Volume rendering is an expensive computation, making training our 3D models under limited memory budgets challenging. However, unlike image-space diffusion models, where the entire image is predicted directly, we can render pixels independent of each other using `render`. In practice, we render $24 \times 24$ patches at random positions in the image in each training iteration. We use the vision transformer architecture from DiT [12] to implement the image backbone `enc` in `denoise`. We modify the MLP architecture (`MLP`) of pixelNeRF to support additional time conditioning input in our models. Our models are trained on 8 A100 GPUs, with a batch size of 24. Training takes around 7 days for RealEstate10k, and around 3 days for Co3D. We use ADAM with a learning rate of $2e-5$. We initially train at a resolution of $64 \times 64$. We then finetune the model at $128 \times 128$. For our Co3D models, we found it helpful to first pretrain the deterministic conditioning component of the model for 10k iterations. We use 64 samples each for coarse and fine stages for volume rendering of the output 3D reconstruction, and only 32 coarse samples for rendering the conditioning input.

We process the Co3D dataset following [5], i.e., we center-crop the images and resize them to a consistent resolution. We follow Chan et al. [5] to provide the absolute pose of the input image as an additional input to the encoder. We also use this input for our baselines, except when using official codebases of SparseFusion [13] and pixelNeRF [14]. During training, we randomly select the initial context frame, and pick a target frame for denoising within predetermined distance intervals. We randomly choose one additional frame between initial context frame and the target frame for computing the novel view reconstruction loss ($\mathcal{L}_{\text{novel}}$). To support autoregressive sampling, we add more context frames from the dataset during training, such that the network can reason jointly from multiple input images. At test time, we iteratively sample new images that are then added as a context frame for the next frame. Autoregressive sampling allows us to cover the entire 360 regions in Co3D scenes by diffusing multiple images around the object. We follow the same training strategy for RealEstate10k, except that we do not feed in absolute poses to our encoder or the baselines. We augment the RealEstate10k dataset by randomly reversing the order of frames in the videos.

**Baselines** We use code provided by the authors for SparseFusion and train on our datasets. We note that the SparseFusion paper only demonstrated results on segmented-out objects without any background, and used multiple input images at test time, unlike our monocular method. Since SparseFusion uses a pretrained VAE backbone that only takes inputs at $256 \times 256$ resolution, we train it at this resolution. However, the 3D optimization is performed at the same resolution as our method. We use the official repository (https://github.com/sxyu/pixel-nerf) for the pixelNeRF baseline. We use 50 scenes for Co3D and 100 scenes for RealEstate for the quantitative evaluations.

## 2.2 Single-Image Motion Prediction

We use an edge-aware smoothness regularization loss on the motion field that is equivalent to the smoothness loss on the depths defined in Sec. 2.1. We use the DiT archicture as our denoising model. The clean context image and the noisy target image are concatenated along the channel dimension and used as input to the network. The output is pixel-aligned motion field that is used to warp the context into the target using SoftSplat [15]. We use ADAM with a learning rate of $2e-5$ and batch size of 72 to optimize our networks on 2 RTXA6000 GPUs. Our models are trained on the Vimeo90K dataset [16].

## 2.3 GAN Inversion

The GAN into which we are inverting is a StyleGAN2-Ada [17] trained on the $256 \times 256$ FFHQ dataset [18]. Our "ground truth" target dataset is generated by taking random samples from this GAN with a truncation $\psi$ of 0.5, down-sampled to $64 \times 64$ pixels. We, again, use the DiT architecture as our denoising model. The context images are obtained by taking a ground truth sample and masking out all but a small patch of varying size. The masked context image and noise target image are concatenated along the channel dimension and used as input to the denoising network, which predicts the denoised "**w**" code. This **w** is then fed through the forward model (generator) and downsampled to

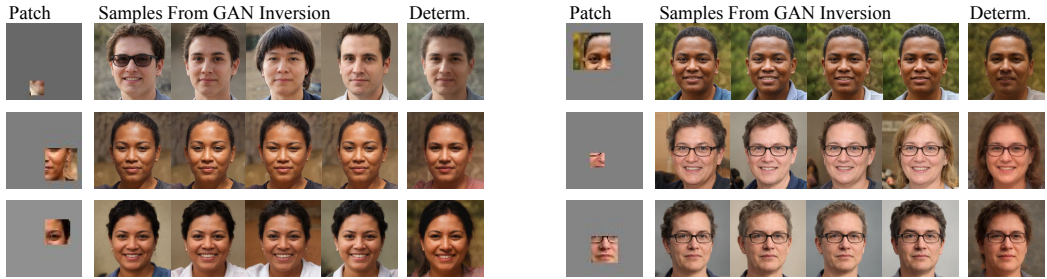

Figure 1: More samples from our GAN Inversion model. Our method produces many plausible faces given only a small patch.

64×64 pixels to obtain our denoised target image. All of our training is done at 64×64 resolution, but using a GAN trained on high-resolution 256×256 images allows us to obtain high-resolution results at test time by simply not downsampling the final denoised output. We use the ADAM optimizer with a learning rate of 2e-5 and a batch size of 4 to train our networks on a single RTXA6000 GPU. We include more results in Figure 1.

### 2.4 Sampling

We use 50 DDIM [19] denoising timesteps for all our results across all applications. All our models, except the inverse graphics model trained on RealEstate10k, are trained without any classifier-free guidance. For our RealEstate10k model, we use a classifier-free guidance weight of 2. Here, the model is also trained as an unconditional model, where the conditioning image is zeroed out for 10% of the iterations.

## 3 Limitations

While we present the first method that enables diffusion models to learn the conditional distribution over signals, only using observations through a forward model, our approach has several limitations. Our sampling times can be very expensive in some cases. The sampling time ranges from just a few seconds for our GAN application, to around 100 mins for 360-degree autoregressive sampling for Co3D. This is both due to the expensive nature of the iterative denoising process, as well as the cost of rendering 3D reconstructions using volume rendering. Our training has large memory requirements, and can thus not be trained on smaller GPUs. Future work on making these models easier to train would make them mode applicable. Our models are not trained on very large-scale datasets, and can thus not generalize to out-of-distribution data. Finally, we only present preliminary investigations into applications outside inverse graphics; however, we hope that we offer a strong experimental base that can be beneficial for future exploration.

# References

[1] Elias M Stein and Rami Shakarchi. *Real analysis*. Princeton lectures in analysis. Princeton University Press, Princeton, NJ, March 2005.

[2] Ben Mildenhall, Pratul P. Srinivasan, Matthew Tancik, Jonathan T. Barron, Ravi Ramamoorthi, and Ren Ng. Nerf: Representing scenes as neural radiance fields for view synthesis. In *Proc. ECCV*, 2020.

[3] Vincent Sitzmann, Semon Rezchikov, Bill Freeman, Josh Tenenbaum, and Fredo Durand. Light field networks: Neural scene representations with single-evaluation rendering. In *NeurIPS*, 2021.

[4] Mehdi SM Sajjadi, Henning Meyer, Etienne Pot, Urs Bergmann, Klaus Greff, Noha Radwan, Suhani Vora, Mario Lučić, Daniel Duckworth, Alexey Dosovitskiy, et al. Scene representation transformer: Geometry-free novel view synthesis through set-latent scene representations. In *CVPR*, 2022.

[5] Eric R Chan, Koki Nagano, Matthew A Chan, Alexander W Bergman, Jeong Joon Park, Axel Levy, Miika Aittala, Shalini De Mello, Tero Karras, and Gordon Wetzstein. Generative novel view synthesis with 3d-aware diffusion models. *arXiv preprint arXiv:2304.02602*, 2023.

[6] Ruoshi Liu, Rundi Wu, Basile Van Hoorick, Pavel Tokmakov, Sergey Zakharov, and Carl Vondrick. Zero-1-to-3: Zero-shot one image to 3d object, 2023.

[7] Calvin Luo. Understanding diffusion models: A unified perspective. *arXiv preprint arXiv:2208.11970*, 2022.

[8] Richard Zhang, Phillip Isola, Alexei A Efros, Eli Shechtman, and Oliver Wang. The unreasonable effectiveness of deep features as a perceptual metric. In *Proc. CVPR*, 2018.

[9] Yang Song, Prafulla Dhariwal, Mark Chen, and Ilya Sutskever. Consistency models. *arXiv preprint arXiv:2303.01469*, 2023.

[10] Sylvain Paris, Pierre Kornprobst, Jack Tumblin, Frédo Durand, et al. Bilateral filtering: Theory and applications. *Foundations and Trends® in Computer Graphics and Vision*, 4(1):1–73, 2009.

[11] Jonathan T Barron, Ben Mildenhall, Dor Verbin, Pratul P Srinivasan, and Peter Hedman. Mip-nerf 360: Unbounded anti-aliased neural radiance fields. In *Proceedings of the IEEE/CVF Conference on Computer Vision and Pattern Recognition*, pages 5470–5479, 2022.

[12] William Peebles and Saining Xie. Scalable diffusion models with transformers. *arXiv preprint arXiv:2212.09748*, 2022.

[13] Zhizhuo Zhou and Shubham Tulsiani. Sparsefusion: Distilling view-conditioned diffusion for 3d reconstruction. *Proc. CVPR*, 2023.

[14] Alex Yu, Vickie Ye, Matthew Tancik, and Angjoo Kanazawa. pixelnerf: Neural radiance fields from one or few images. In *Proc. CVPR*, 2021.

[15] Simon Niklaus and Feng Liu. Softmax splatting for video frame interpolation. In *Proc. CVPR*, 2020.

[16] Tianfan Xue, Baian Chen, Jiajun Wu, Donglai Wei, and William T Freeman. Video enhancement with task-oriented flow. *International Journal of Computer Vision*, 127:1106–1125, 2019.

[17] Tero Karras, Miika Aittala, Janne Hellsten, Samuli Laine, Jaakko Lehtinen, and Timo Aila. Training generative adversarial networks with limited data. *Advances in neural information processing systems*, 33:12104–12114, 2020.

[18] Tero Karras, Samuli Laine, and Timo Aila. A style-based generator architecture for generative adversarial networks. In *Proc. CVPR*, 2019.

[19] Jiaming Song, Chenlin Meng, and Stefano Ermon. Denoising diffusion implicit models. In *ICLR*, 2021.


\end{document}